\documentclass[acmsmall]{acmart}
\usepackage{booktabs}
\usepackage{epsfig}
\usepackage{graphicx}
\usepackage{amsmath}
\usepackage{amssymb}
\usepackage{bm}

\usepackage{algorithm}
\usepackage{algorithmic}
\usepackage{url}
\usepackage{subfigure}
\usepackage{bbding}
\usepackage{multirow}
\usepackage{enumitem}
\usepackage{array}
\def\BibTeX{{\rm B\kern-.05em{\sc i\kern-.025em b}\kern-.08emT\kern-.1667em\lower.7ex\hbox{E}\kern-.125emX}}
    
\begin{document}

%
\title{Transfer Learning with Dynamic Distribution Adaptation}

%
\author{Jindong Wang}

\email{jindong.wang@microsoft.com}
\affiliation{%
	\institution{Microsoft Research Asia}
	\streetaddress{No. 5 Danling Street}
	\city{Beijing}
	\postcode{100080}
	\country{China}
}

\author{Yiqiang Chen}
\authornote{Corresponding author: Yiqiang Chen.}
\email{yqchen@ict.ac.cn}
\author{Wenjie Feng}
\email{fengwenjie@ict.ac.cn}
\affiliation{%
  \institution{Institute of Computing Technology, Chinese Academy of Sciences}
  \streetaddress{No. 6 Kexueyuan South Road}
  \city{Beijing}
  \postcode{100190}
  \country{China}
}

\author{Han Yu}
\affiliation{%
  \institution{School of Computer Science and Engineering, Nanyang Technological University}
  \country{Singapore}}
\email{han.yu@ntu.edu.sg}

\author{Meiyu Huang}
\email{huangmeiyu@qxslab.cn}
\affiliation{%
  \institution{Qian Xuesen Laboratory of Space Technology, China Academy of Space Technology}
  \city{Beijing}
  \country{China}
}

\author{Qiang Yang}
\email{qyang@cse.ust.hk}
\affiliation{%
 \institution{Department of Computer Science and Engineering, Hong Kong University of Science and Technology}
 \city{Kowloon}
 \country{Hong Kong}
}

%
\renewcommand{\shortauthors}{Wang, et al.}

%
\begin{abstract}
Transfer learning aims to learn robust classifiers for the target domain by leveraging knowledge from a source domain. Since the source and the target domains are usually from different distributions, existing methods mainly focus on adapting the cross-domain marginal or conditional distributions. However, in real applications, the marginal and conditional distributions usually have different contributions to the domain discrepancy. Existing methods fail to quantitatively evaluate the different importance of these two distributions, which will result in unsatisfactory transfer performance. In this paper, we propose a novel concept called Dynamic Distribution Adaptation (DDA), which is capable of quantitatively evaluating the relative importance of each distribution. DDA can be easily incorporated into the framework of structural risk minimization to solve transfer learning problems. On the basis of DDA, we propose two novel learning algorithms: (1) Manifold Dynamic Distribution Adaptation (MDDA) for traditional transfer learning, and (2) Dynamic Distribution Adaptation Network (DDAN) for deep transfer learning. Extensive experiments demonstrate that MDDA and DDAN significantly improve the transfer learning performance and setup a strong baseline over the latest deep and adversarial methods on digits recognition, sentiment analysis, and image classification. More importantly, it is shown that marginal and conditional distributions have different contributions to the domain divergence, and our DDA is able to provide good quantitative evaluation of their relative importance which leads to better performance. We believe this observation can be helpful for future research in transfer learning.
\end{abstract}

%
%


\setcopyright{acmcopyright}
\acmJournal{TIST}
\acmYear{2019} \acmVolume{1} \acmNumber{1} \acmArticle{1} \acmMonth{1} \acmPrice{15.00}\acmDOI{10.1145/3360309}

%

%

%
\maketitle

\section{Introduction}
Supervised learning is perhaps the most popular and well-studied paradigm in machine learning during the past years. Significant advances have been achieved in supervised learning by exploiting a large amount of \textit{labeled} training data to build powerful models. For instance, in computer vision, large-scale labeled datasets such as ImageNet~\cite{deng2009imagenet} for image classification and MS COCO~\cite{lin2014microsoft} for object detection and semantic segmentation have played an instrumental role to help train computer vision models with superior performance. In sentiment analysis, a lot of reviews for all kinds of products are available to train sentiment classification models. Unfortunately, it is often expensive and time-consuming to acquire sufficient labeled data to train these models. Furthermore, there is often \textit{dataset bias} in newly emerging data, i.e. the existing model is often trained on a particular dataset and will generalize poorly in a new domain. For example, the images of an online product can be very different from those taken at home. The product review for electronic devices is likely to be different from that of clothes. Under this circumstance, it is necessary and important to design algorithms that can handle both the label scarcity and dataset bias challenges.

Domain adaptation, or transfer learning~\cite{pan2010survey,transferlearning} has been a promising approach to solve such problems. The main idea of transfer learning is to leverage the abundant labeled samples in some existing domains to facilitate learning in a new target domain by reducing the dataset bias. The domain with abundant labeled samples is often called the \textit{source} domain, while the domain for which a new model is to be trained is the \textit{target} domain. However, due to the dataset bias, the data distributions on different domains are usually different. In such circumstance, traditional machine learning algorithms cannot be applied directly since they assume that training and testing data are under the same distributions. Transfer learning is able to reduce the distribution divergence~\cite{pan2010survey} such that the models on the target domain can be learned.

To cope with the difference in distributions between domains, existing works can be summarized into two main categories: (a)~\textit{instance reweighting}~\cite{dai2007boosting,xu2017unified}, which reuses samples from the source domain according to some weighting technique; and (b)~\textit{feature matching}, which either performs subspace learning by exploiting the subspace geometrical structure~\cite{wang2019easy,fernando2013unsupervised,sun2016return,gong2012geodesic}, or distribution alignment to reduce the marginal or conditional distribution divergence between domains~\cite{wang2017balanced,zhang2017joint,long2013transfer}. Recently, the success of deep learning has dramatically increased the performance of transfer learning either via deep representation learning~\cite{zhu2019multi,yosinski2014transferable,he2016deep,bousmalis2016domain,long2017deep,wang2018deep} or adversarial learning~\cite{ganin2016domain,ganin2014unsupervised,long2018conditional,zhang2018collaborative,sankaranarayanan2017generate}. The key idea behind these works is to learn more transferable representations using deep neural networks. Then, the learned feature distributions can be aligned such that their domain discrepancy can be reduced.

\begin{figure}[t!]
	\centering
	\subfigure[Different target distributions]{
		\includegraphics[scale=0.35]{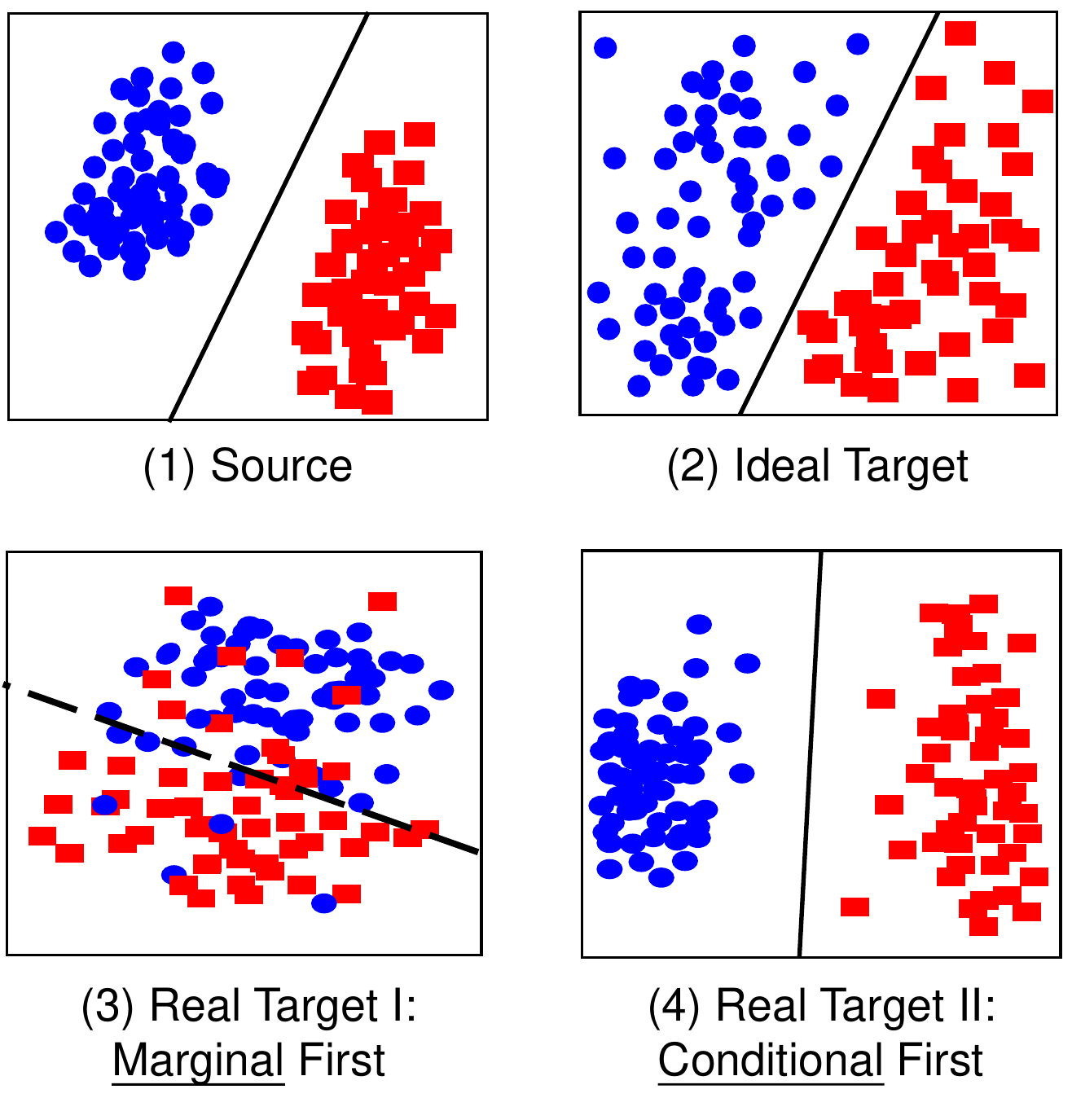}
		\label{fig-sub-sourcetarget}}
	\hspace{.2in}
	\subfigure[Performance of our DDA and the latest methods]{
		\includegraphics[scale=0.47]{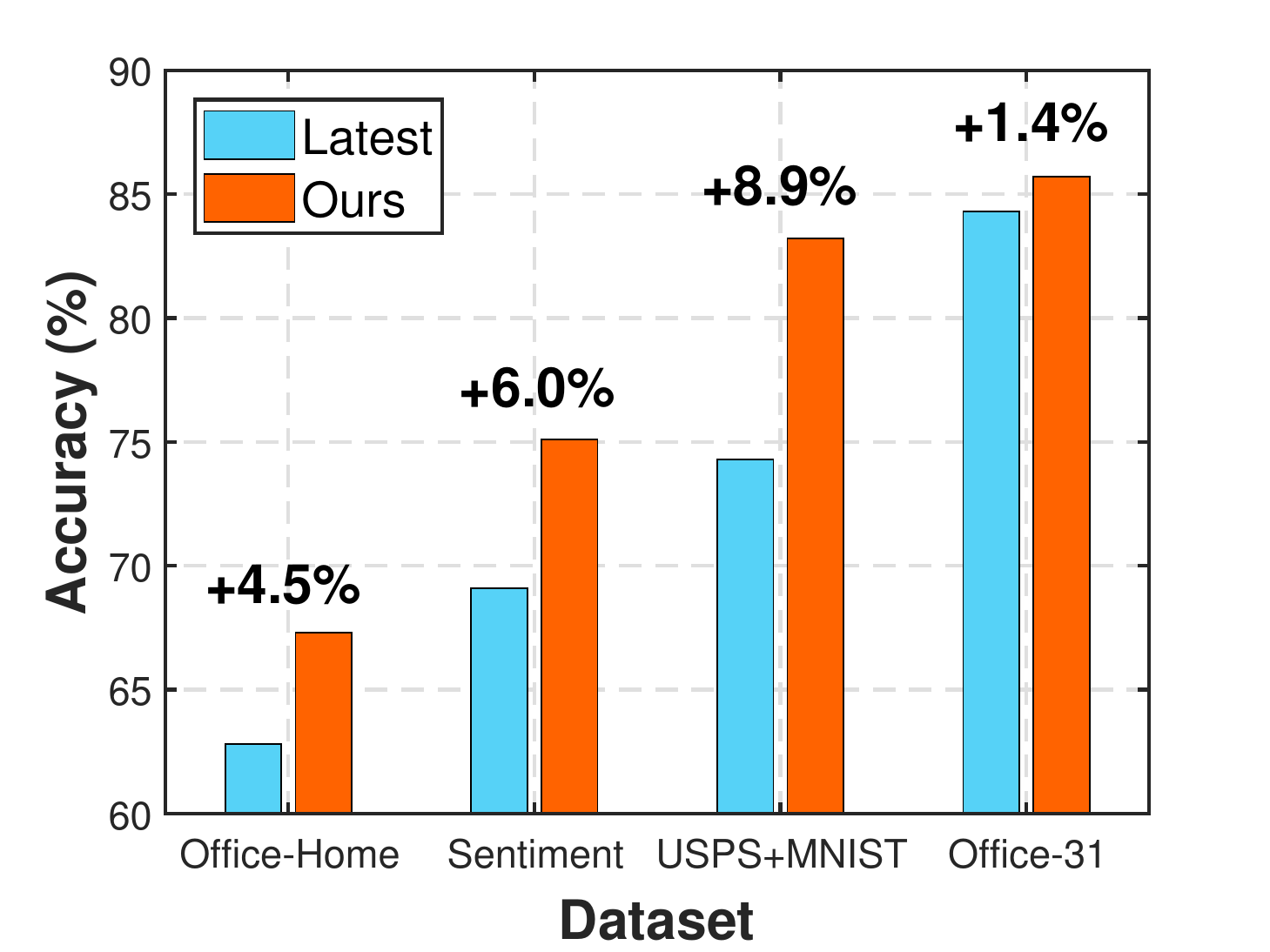}
		\label{fig-sub-res}}
	
	\caption{(a) The different effect of marginal and conditional distributions. (b) Performance comparison between our dynamic distribution adaptation and the latest transfer learning methods.}
	\label{fig-intro}
\end{figure}

However, despite the great success achieved by traditional and deep transfer learning methods, there is still a challenge ahead. Existing works only attempt to align the marginal~\cite{pan2011domain,tzeng2014deep,long2015learning} or the conditional distributions~\cite{pei2018multi,long2018conditional}. Although recent advance has suggested that aligning both distributions will lead to better performance~\cite{long2013transfer,wang2017balanced,long2017deep}, they only give the two distributions \textit{equal} weights, which fails to evaluate the relative importance of these two distributions. In their assumptions, both the marginal and conditional distributions are contributing equally to the domain divergence. However, in this paper, we argue that this assumption is not practical in real applications. For example, when two domains are very dissimilar~(e.g., transfer learning between (1) and (3) in Fig.~\ref{fig-sub-sourcetarget}), the marginal distribution is more important. When the marginal distributions are close~(transfer learning between (1) and (4) in Fig.~\ref{fig-sub-sourcetarget}), the conditional distribution of each class should be given more weight. Ignoring this fact will likely to result in unsatisfactory transfer performance. There is no method which can \textit{quantitatively} account for the relative importance of these two distributions in conjunction.

In this paper, we propose a novel concept of \textbf{Dynamic Distribution Adaptation (DDA)} to dynamically and quantitatively adapt the marginal and conditional distributions in transfer learning. To be specific, DDA is able to dynamically learn the distribution weights through calculating the $\mathcal{H} \Delta \mathcal{H}$ divergence~\cite{ben2007analysis} between domains when learning representations. Then, the relative importance of marginal and conditional distributions can be obtained, which in turn can be utilized to learn more transferable feature representations. This dynamic importance learning and feature learning are being optimized iteratively to learn a domain-invariant transfer classifier eventually. To the best of our knowledge, DDA is the first work to dynamically and quantitatively evaluate the importance of both distributions. The significant improvements of DDA over the latest method on different kinds of tasks are shown in Fig.~\ref{fig-sub-res}.

To enable good representation learning, we propose two novel learning methods based on DDA with the principle of Structural Risk Minimization (SRM)~\cite{vapnik1998statistical}. For traditional transfer learning, we develop \textit{Manifold Dynamic Distribution Adaptation (MDDA)} method to utilize the Grassmann manifold~\cite{hamm2008grassmann} in learning non-distorted feature representations. For deep transfer learning, we develop \textit{Dynamic Distribution Adaptation Network (DDAN)} to use the deep neural work in learning end-to-end transfer classifier. We also develop respective learning algorithms for MDDA and DDAN. 

To sum up, this work makes the following contributions:

1) We propose the DDA concept for domain adaptation. DDA is the \textit{first} quantitative evaluation framework for the relative importance of marginal and conditional distributions in domain adaptation. This is useful for a wide range of future research on transfer learning.

2) On top of DDA, we propose two novel methods: MDDA for traditional transfer learning and DDAN for deep transfer learning. Both methods can be efficiently formulated and finally learn the domain-invariant transfer classifier.

3) We conduct extensive experiments on digit classification, object recognition, and sentiment classification datasets. Experimental results demonstrate that both MDDA and DDAN are significantly better than many state-of-the-art traditional and deep methods. More importantly, empirical results have also demonstrated that the different effect of marginal and conditional distributions do exist, and our DDA is able to give them quantitative weights, which facilitates the performance of transfer learning.

This paper is an extension of our previous oral paper at ACM Multimedia conference\cite{wang2018visual}. Our extensions include: (1) A more general and clear concept of dynamic distribution adaptation and its calculation. (2) We extend DDA in both manifold learning and deep learning methods, then we formulate these algorithms and propose respective learning algorithms. (3) We extend the experiments in digit classification, sentiment analysis, and image classification, which have shown the effectiveness of our methods. And (4) We extensively analyze our calculation of DDA in new experiments.

The remainder of this paper is structured as follows. We review the related work in Section~\ref{sec-related}. In Section~\ref{sec-pre}, we introduce some previous knowledge before introducing the proposed method. Section~\ref{sec-method} thoroughly presents our proposed DDA concept and its two extensions: MDDA and DDAN. Extensive experiments are shown in Section~\ref{sec-exp}, where we extensively evaluate the performance of MDDA and DDAN. Finally, Section~\ref{sec-con} concludes this paper.

\section{Related Work}
\label{sec-related}

Domain adaptation, or transfer learning, is an active research area in machine learning. Apart from the popular survey by Pan and Yang~\cite{pan2010survey}, several recent survey papers have extensively investigated specific research topics in transfer learning including: visual domain adaptation~\cite{wang2018visual,wang2019easy}, heterogeneous transfer learning~\cite{day2017survey,friedjungova2017asymmetric}, multi-task learning~\cite{zhang2017survey}, and cross-dataset recognition~\cite{zhang2017cross}. There are also several successful applications using transfer learning for: activity recognition~\cite{wang2018stratified,wang2018deep,chen2019fedhealth,chen2019cross}, object recognition~\cite{ghifary2017scatter}, face recognition~\cite{ren2014transfer}, speech recognition~\cite{yi2019language}, speech synthesis~\cite{jia2018transfer}, and text classification~\cite{do2006transfer}. Interested readers are recommended to refer to \url{http://transferlearning.xyz} to find out more related works and applications.

From the perspective of transfer learning methods, there are three main categories: (1) Instance re-weighting, which reuses samples according to some weighting technique~\cite{dai2007boosting,dai2007co,tan2015transitive}; (2) Feature transformation, which performs representation learning to transform the source and target domains into the same subspace~\cite{long2018conditional,wang2017balanced,zhang2017joint,pan2011domain}; (3) Transfer metric learning~\cite{luo2018transfer,luo2014decomposition,luo2018online}, which learns transferable metric between domains. Since our proposed methods are mainly related to feature-based transfer learning, we will extensively introduce the related work in the following aspects.

%

\subsection{Subspace and Manifold Learning}

One category of feature-based transfer learning is subspace and manifold learning. The goal is to learn representative subspace or manifold representations that are invariant across domains. Along this line, subspace alignment~(SA)~\cite{fernando2013unsupervised} aligned the base vectors of both domains, but failed to adapt feature distributions. Subspace distribution alignment (SDA)~\cite{sun2015subspace} extended SA by adding subspace variance adaptation. However, SDA did not consider the local property of subspaces and ignored conditional distribution alignment. CORAL~\cite{sun2016return} aligned subspaces in second-order statistics, but it did not consider the distribution alignment. Scatter component analysis~(SCA)~\cite{ghifary2017scatter} converted the samples into a set of subspaces ~(i.e. scatters) and then minimized the divergence between them. 

On the other hand, some work used the property of manifold to further learn tight representations. Geodesic flow kernel~(GFK)~\cite{gong2012geodesic} extended the idea of sampled points in manifold~\cite{gopalan2011domain} and proposed to learn the geodesic flow kernel between domains. The work of \cite{baktashmotlagh2014domain} used a Hellinger distance to approximate the geodesic distance in Riemann space. \cite{baktashmotlagh2013unsupervised} proposed to use Grassmann for domain adaptation, but they ignored the conditional distribution alignment. Different from these approaches, DDA can learn a domain-invariant classifier in the manifold and align both marginal and conditional distributions.

\subsection{Distribution Alignment}

Another category of feature-based transfer learning is distribution alignment. The work of this category is pretty straight forward: find some feature transformations that can minimize the distribution divergence. Along this line, existing work can be classified into three subcategories: marginal distribution alignment, conditional distribution alignment, and joint distribution alignment. 

DDA substantially differs from existing work that only aligns marginal or conditional distribution~\cite{pan2011domain}. Joint distribution adaptation~(JDA)~\cite{long2013transfer} matched both distributions with equal weights. Others extended JDA by adding regularization~\cite{long2014adaptation}, sparse representation~\cite{xu2016discriminative}, structural consistency~\cite{hou2016unsupervised}, domain invariant clustering~\cite{tahmoresnezhad2016visual}, and label propagation~\cite{zhang2017joint}. The work of Balanced Distribution Adaptation (BDA)~\cite{wang2017balanced} firstly proposed to manually weight the two distributions. The main differences between DDA~(MDDA) and these methods are: 1) These work treats the two distributions equally. However, when there is a greater discrepancy between both distributions, they cannot evaluate their relative importance and thus lead to undermined performance. Our work is capable of evaluating the quantitative importance of each distribution via considering their different effects. 2) These methods are designed only for the original space, where feature distortion will negatively affect the performance. DDA~(MDDA) can align the distributions in the manifold to overcome the feature distortion.

\subsection{Domain-invariant Classifier Learning}

Different from the above two types of work that further need to learn a classifier for the target domain, some research is able to learn the domain-invariant classifier while simultaneously performing subspace learning or distribution alignment. Recent work such as adaptation regularization for transfer learning (ARTL)~\cite{long2014adaptation}, domain-invariant projection (DIP)~\cite{baktashmotlagh2013unsupervised,baktashmotlagh2016distribution}, and distribution matching machines (DMM)~\cite{cao2018unsupervised} also aimed to build a domain-invariant classifier. However, ARTL and DMM cannot effectively handle feature distortion in the original space. Nor can they account for the different importance of distributions. DIP mainly focused on feature transformation and only aligned marginal distributions. DDA~(MDDA) is able to mitigate feature distortion and quantitatively evaluate the importance of marginal and conditional distribution adaptation.

\subsection{Deep and Adversarial Transfer Learning}

Recent years have witnessed the advance of deep transfer learning. Compared to traditional shallow learning, deep neural networks are capable of learning better representations~\cite{yosinski2014transferable}. Deep domain confusion~(DDC)~\cite{tzeng2014deep} firstly added the MMD loss to a deep network to adapt the network. Similar to DDC, deep adaptation networks~(DAN) adopted the multiple-kernel MMD~\cite{gretton2012optimal} to the network. Instead, Deep CORAL~\cite{sun2016deep} added CORAL loss~\cite{sun2016return} to the network. CORAL is a second-order loss compared to MMD, which is a first-order loss. Furthermore, Zellinger \textit{et al.} introduce the central moment discrepancy~(CMD)~\cite{zellinger2017central} to the network, which is a higher-order distance. 

Different from the above deep transfer learning methods, adversarial learning~\cite{goodfellow2014generative} also helps to learn more transferable and discriminative representations. Domain-adversarial neural network (DANN) was first introduced by Ganin \textit{et al.}~\cite{ganin2014unsupervised,ganin2016domain}. The core idea is to add a domain-adversarial loss to the network instead of the predefined distance function such as MMD. This has dramatically enabled the network to learn more discriminative information. Following the idea of DANN, a lot of work adopted domain-adversarial training~\cite{long2017deep,zhang2018collaborative,long2018conditional,bousmalis2016domain,chen2018joint}.

The above-discussed work all ignore the different effect of marginal and conditional distributions in transfer learning, while our proposed DDA (DDAN) is fully capable of dynamically evaluating the importance of each distribution.

\section{Preliminaries}
\label{sec-pre}

Let $\Omega \in \mathbb{R}^d$ be an input measurable space of dimension $d$ and $\mathcal{C}$ the set of possible labels. We use $P(\Omega)$ to denote the set of all probability measures over $\Omega$. In standard transfer learning setting, there is a source domain $\Omega_s = \{\mathbf{x}^s_i, y^s_i\}^{n_s}_{i=1}$ with known labels $y^s_i \in \mathcal{C}$ and a target domain $\Omega_t = \{\mathbf{x}^t_j\}^{n_t}_{j=1}$ with unknown labels. Here, $\mathbf{x}^s \sim P(\Omega_s)$ and $\mathbf{x}^t \sim P(\Omega_t)$ are samples from either source or the target domain. Different from existing work that either assume the marginal or conditional distributions of two domains are different, in this work, we tackle a more general case that \textit{both} distributions are different, i.e. $P(\mathbf{x}^s) \ne P(\mathbf{x}^t), P(y^s|\mathbf{x}^s) \ne P(y^t|\mathbf{x}^t)$. The goal is to learn a transferable classifier $f$ such that the risk on the target domain can be minimized: $\epsilon_t=\min P_{(\mathbf{x},y) \sim \Omega_t}(f(\mathbf{x}) \ne y)$.

\subsection{Structural Risk Minimization}
From a statistical machine learning perspective, the above problem can be formulated and solved by the structural risk minimization~(SRM) principle~\cite{vapnik1998statistical}. In SRM, the prediction function $f$ can be formulated as
\begin{equation}
\label{eq-srm}
	f = \mathop{\arg\min}_{f \in \mathcal{H}_{K}, (\mathbf{x},y) \sim \Omega_l} J(f(\mathbf{x}),y) + \lambda R(f),
\end{equation}
where the first term indicates the loss on data samples with $J(\cdot,\cdot)$ is the loss function, the second term denotes the regularization term, and $\mathcal{H}_{K}$ is the Hilbert space induced by kernel function $K(\cdot,\cdot)$. $\lambda$ is the trade-off parameter. The symbol $\Omega_l$ denotes the domain that has labels. 

In our problem, we have $\Omega_l = \Omega_s$ since there are no labels in the target domain. Specifically, in order to effectively handle the different distributions between $\Omega_s$ and $\Omega_t$, we can further divide the regularization term as
\begin{equation}
	R(f) = \lambda \overline{D_f}(\Omega_s,\Omega_t) + \rho R_f(\Omega_s,\Omega_t),
\end{equation}
where $\overline{D_f}(\cdot, \cdot)$ represents the distribution divergence between $\Omega_s$ and $\Omega_t$ with $\lambda,\rho$ the trade-off parameters and $R_f(\cdot, \cdot)$ denotes other regularization.

\subsection{Maximum Mean Discrepancy}
There are a variety of means to measure the distribution divergence between two domains such as Kullback$-$Leibler divergence and cross-entropy. With respect to efficiency, we adopt the \textit{maximum mean discrepancy} (MMD)~\cite{ben2007analysis} to empirically calculate the distribution divergence between domains. As a non-parametric measurement, MMD has been widely adopted by many existing methods~\cite{zhang2017joint,ghifary2017scatter,pan2011domain}, and its effectiveness has been proven analytically in~\cite{gretton2012kernel}. 

Formally, the MMD distance between distributions $P$ and $Q$ is defined as~\cite{gretton2012kernel}
\begin{equation}
	MMD(\mathcal{H}_k, P, Q) := \sup_{||f||_{\mathcal{H}_k} \le 1} \mathbb{E}_{X \sim P} f(X) - \mathbb{E}_{Y \sim Q} f(Y),
\end{equation}
where $\mathcal{H}_k$ is the Reproduced kernel Hilbert space (RKHS) with Mercer kernel $K(\cdot,\cdot)$, $||f||_{\mathcal{H}_k} \le 1$ is its unit norm ball, and $\mathbb{E}[\cdot]$ denotes the mean of the embedded samples.

This is known as an integral probability metric in the statistics literature. To compute this divergence, a biased empirical estimate of MMD is obtained by replacing the population expectations with empirical expectations computed on the samples $X$ and $Y$,
\begin{equation}
\label{eq-mmd}
	MMD_b(\mathcal{H}_k, P, Q) = \sup_{||f||_{\mathcal{H}_k} \le 1} \left( \frac{1}{m} \sum_{i=1}^{m} f(X_i) - \frac{1}{n} \sum_{i=1}^{n} f(Y_i) \right),
\end{equation}
where $m,n$ are sample numbers of $P$ and $Q$, respectively.

\section{Dynamic Distribution Adaptation}
\label{sec-method}

In this section, we present the general dynamic distribution adaptation framework and its two learning algorithms in detail.

\subsection{The General Framework}
Transfer learning is to learn transferable representations which can generalize well across different domains. The key idea of Dynamic Distribution Adaptation~(DDA) is to dynamically learn the relative importance of marginal and conditional distributions in transfer learning. Therefore, the dynamic importance learning and transfer feature learning are not independently, but quite involved. Accordingly, DDA first performs feature learning to learn more transferable representations. Then, it can perform dynamic distribution adaptation to quantitatively account for the relative importance of marginal and conditional distributions to address the challenge of unevaluated distribution alignment. These two steps are iteratively optimized via several iterations. Eventually, a domain-invariant classifier $f$ can be learned by combining these two steps based on the principle of SRM. 

Recall the principle of SRM in Eq.~(\ref{eq-srm}). If we use $g(\cdot)$ to denote the feature learning function, then $f$ can be represented as

\begin{equation}
\label{equ-f-orig}
f = \mathop{\arg\min}_{f \in \sum_{i=1}^{n} \mathcal{H}_{K}} J(f(g(\mathbf{x}_i)),y_i) + \eta ||f||^2_K + \lambda \overline{D_f}(\Omega_s,\Omega_t) + \rho R_f(\Omega_s,\Omega_t)
\end{equation}
where $||f||^2_K$ is the squared norm of $f$. The term $\overline{D_f}(\cdot,\cdot)$ represents the proposed dynamic distribution alignment. Additionally, we introduce $R_f(\cdot,\cdot)$ as a Laplacian regularization to further exploit the similar geometrical property of nearest points in manifold $\mathbb{G}$~\cite{belkin2006manifold}. $\eta,\lambda$, and $\rho$ are the regularization parameters.

\begin{figure}[t!]
	\centering
	\subfigure[MDDA]{
		\includegraphics[scale=0.54]{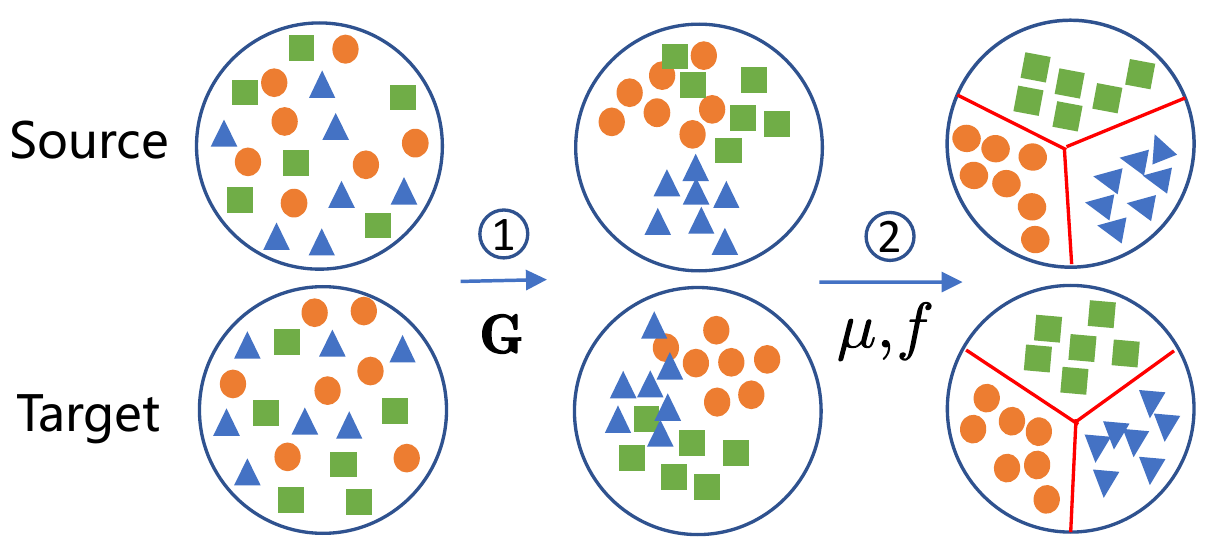}
		\label{fig-sub-manifold}}
	\hspace{.3in}
	\subfigure[DDAN]{
		\includegraphics[scale=0.52]{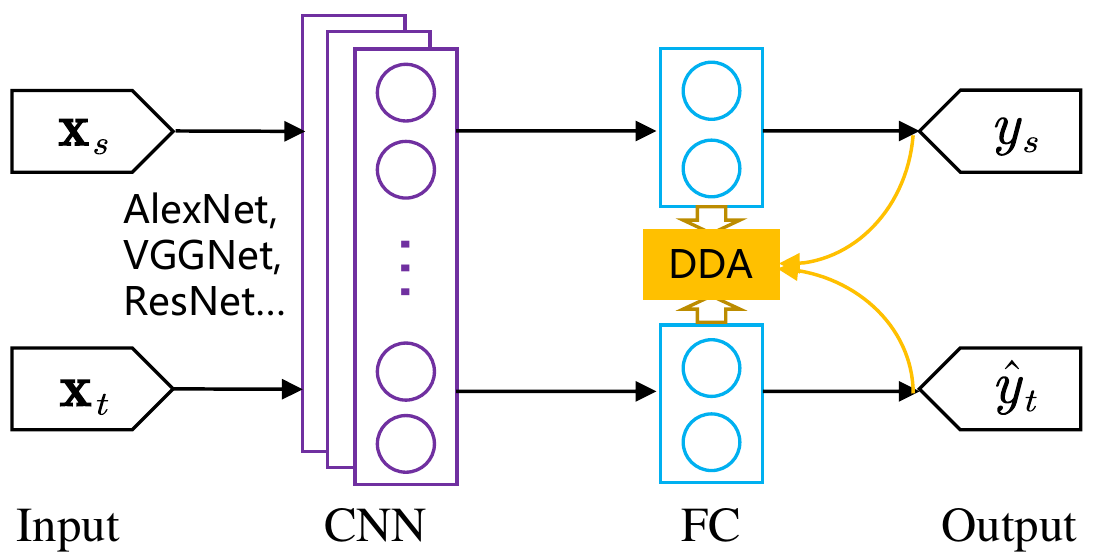}
		\label{fig-sub-deep}}
	
	\caption{The main idea of MDDA (Manifold Dynamic Distribution Adaptation) and DDAN~(Dynamic Distribution Adaptation Network)}
	\label{fig-main}
\end{figure}

In the next sections, we first introduce the learning of dynamic distribution adaptation. Then, we show how to learn the feature learning function $g(\cdot)$ either through manifold learning (i.e. Manifold Dynamic Distribution Adaptation, or MDDA in Fig.~\ref{fig-sub-manifold}) and deep learning (i.e. Dynamic Distribution Network, or DDAN in Fig.~\ref{fig-sub-deep}).


\subsection{Dynamic Distribution Adaptation}
\label{sec-da}

The purpose of dynamic distribution adaptation is to \textit{quantitatively} evaluate the importance of aligning marginal~($P$) and conditional~($Q$) distributions in domain adaptation.
Existing methods~\cite{long2013transfer,zhang2017joint} assume that both distributions are equally important. However, this assumption may not be valid in real-world applications. For instance, when transferring from (1) to (3) in Fig.~\ref{fig-sub-sourcetarget}, there is a large difference between datasets. Therefore, the divergence between $P_s$ and $P_t$ is more dominant. In contrast, from (1) to (4) in Fig.~\ref{fig-sub-sourcetarget}, the datasets are similar. Therefore, the distribution divergence in each class ($Q_s$ and $Q_t$) is more dominant. 

In view of this phenomenon, we introduce an \textit{adaptive factor} to dynamically adjust the importance of these two distributions. Formally, the dynamic distribution alignment $\overline{D_f}$ is defined as
\begin{equation}
\label{equ-bda}
\overline{D_{f}}(\Omega_s,\Omega_t) = (1- \mu)D_{f}(P_s,P_t) + \mu \sum_{c=1}^{C} D^{(c)}_{f}(Q_s,Q_t)
\end{equation}
where $\mu \in [0,1]$ is the adaptive factor and $c \in \{1,\cdots,C\}$ is the class indicator. $D_f(P_s,P_t)$ denotes the marginal distribution alignment, and $D^{(c)}_f(Q_s,Q_t)$ denotes the conditional distribution alignment for class $c$.

When $\mu \rightarrow 0$, it means that the distribution distance between the source and the target domains is large. Thus, marginal distribution alignment is more important~((1) $\rightarrow$ (3) in Fig.~\ref{fig-sub-sourcetarget}). When $\mu \rightarrow 1$, it means that feature distribution between domains is relatively small, such that the distribution of each class is dominant. Thus, the conditional distribution alignment is more important ((1) $\rightarrow$ (4) in Fig.~\ref{fig-sub-sourcetarget}). When $\mu=0.5$, both distributions are treated equally as in existing methods~\cite{long2013transfer,zhang2017joint}. Hence, the existing methods can be regarded as special cases of the dynamic distribution alignment. By learning the optimal adaptive factor $\mu_{opt}$ (which we will discuss later), MDDA can be applied to different domain adaptation problems.

We use the \textit{maximum mean discrepancy} (MMD)~\cite{ben2007analysis} introduced in the last section to empirically calculate the distribution divergence between domains. To be specific, the marginal and conditional distribution distances can be respectively computed as
\begin{equation}
	D_f(P_s,P_t)=\Vert\mathbb{E}[f(\mathbf{z}_s)] - \mathbb{E}[f(\mathbf{z}_t)]\Vert^2_{\mathcal{H}_{K}}
\end{equation}
\begin{equation}
	D^{(c)}_f(Q_s,Q_t)=\Vert \mathbb{E}[f(\mathbf{z}^{(c)}_s)] - \mathbb{E}[f(\mathbf{z}^{(c)}_t)]\Vert^2_{\mathcal{H}_{K}}
\end{equation}

Then, DDA can be expressed as
\begin{equation}
\label{equ-da1}
\overline{D_{f}}(\Omega_s,\Omega_t) = (1 - \mu) \Vert\mathbb{E}[f(\mathbf{z}_s)) - \mathbb{E}[f(\mathbf{z}_t)]\Vert^2_{\mathcal{H}_{K}} + \mu \sum_{c=1}^{C}\Vert \mathbb{E}[f(\mathbf{z}^{(c)}_s)] - \mathbb{E}[f(\mathbf{z}^{(c)}_t)]\Vert^2_{\mathcal{H}_{K}}.
\end{equation}

Note that since $\Omega_t$ has no labels, it is not feasible to evaluate the conditional distribution $Q_t=Q_t(y_t|\mathbf{z}_t)$. Instead, we follow the idea in~\cite{wang2017balanced} and use the class conditional distribution $Q_t(\mathbf{z}_t|y_t)$ to approximate $Q_t$. In order to evaluate $Q_t(\mathbf{z}_t|y_t)$, we apply prediction to $\Omega_t$ using a base classifier trained on $\Omega_s$ to obtain soft labels for $\Omega_t$. The soft labels may be less reliable, so we \textit{iteratively} refine the prediction. Note that we \textit{only} use the base classifier in the first iteration. After that, MDDA can \textit{automatically} refine the labels for $\Omega_t$ using results from previous iterations.

\subsubsection{Quantitative Evaluation of Adaptive Factor $\bm{\mu}$}

We can treat $\mu$ as a parameter and tune its value by cross-validation techniques. However, there are no labels for the target domain in unsupervised transfer learning problems. There are two indirect solutions to apply the value of $\mu$ in DDA rather than estimating its value: by \textit{Random guessing} and by \textit{Max-min averaging}. Random guessing is technically very intuitive. We can randomly pick a value of $\mu$ in $[0,1]$, then perform MDDA using the corresponding $\mu_{rand}$ to get the transfer learning result. If we repeat this process $t$ times and denote the $t$-th transfer learning result as $r_t$, then the final result can be calculated as $\frac{1}{t} \sum_{i=1}^{t} r_t$. Max-min averaging is also simple to implement. We can search the value of $\mu$ in $[0,1]$ with the step of 0.1, which will generate a candidate set of $\mu$: $[0,0.1,\cdots,0.9,1.0]$. Then, similar to random guessing, we can also obtain the averaged results as $\frac{1}{11} \sum_{i=1}^{11} r_i$.

Although the random guessing and max-min averaging are both feasible and simple solutions to estimate $\mu$, they are computationally prohibitive. More importantly, there is no guarantee of their results. It is extremely challenging to calculate the value of $\mu$.

In this work, we make the \textit{first} attempt towards calculating $\mu$ (i.e. $\hat{\mu}$) by exploiting the global and local structure of domains. We adopt the $\mathcal{A}$-distance~\cite{ben2007analysis} as the basic measurement. The $\mathcal{A}$-distance is defined as the error of building a linear classifier to distinguish two domains (i.e. a binary classification). Formally, we denote $\epsilon(h)$ the error of a linear classifier $h$ discriminating the two domains $\Omega_s$ and $\Omega_t$. Then, the $\mathcal{A}$-distance can be defined as
\begin{equation}
d_A(\Omega_s,\Omega_t) = 2(1 - 2 \epsilon(h)).
\end{equation}

We can directly compute the marginal $\mathcal{A}$-distance using the above equation, which is denoted as $d_M$. For the $\mathcal{A}$-distance between conditional distributions, we use $d_c$ to denote the $\mathcal{A}$-distance for the $c$th class. It can be calculated as $d_c = d_A(\mathcal{D}^{(c)}_s,\mathcal{D}^{(c)}_t)$, where $\mathcal{D}^{(c)}_s$ and $\mathcal{D}^{(c)}_t$ denote samples from class $c$ in $\Omega_s$ and $\Omega_t$, respectively. Note that $d_M$ denotes the marginal difference, while $sum_{c=1}^{C} d_c$ denotes the conditional difference on all classes. In this paper, our assumption is that the domain divergence is caused by both the marginal and conditional distributions. Therefore, $d_M + \sum_{c=1}^{C} d_c$ can represent the whole divergence. Eventually, $\mu$ can be estimated as

\begin{equation}
\label{eq-mu}
\hat{\mu} = 1 - \frac{d_M}{d_M + \sum_{c=1}^{C} d_c}.
\end{equation}

Note that the number of labeled samples in the source domain is often much larger than that in the target domain. Therefore, in order to solve this imbalanced classification problem, we perform upsampling~\cite{krawczyk2016learning} on the target domain to make the samples with almost the same size. We also notice that this upsampling process is random, thus we repeat this step several times to get the averaged $\mu$ value.

This estimation has to be computed during every iteration of the dynamic distribution adaptation, since the feature distribution may vary after evaluating the conditional distribution each time. 
To the best of our knowledge, this is the \textit{first} solution to quantitatively estimate the relative importance of each distribution. In fact, this estimation can be significant for future research in transfer learning and domain adaptation.

\textit{Remark:} Currently, the quantitative evaluation of $\mu$ only supports the situation where the label space of the source and the target domains are the same. However, it is important to note that this evaluation is also open for open set domain adaptation~\cite{panareda2017open} or partial transfer learning~\cite{zhang2018importance}, where the label spaces are not the same. In such cases, we should consider the different similarity between each class in two domains. For instance, we can regard the classes that do not belong to the target domain as outliers and perform outlier detection before calculating $\mu$. Then, we can select the most similar samples and classes in both domains and perform DDA. We leave this part for future research.

\subsection{MDDA: Manifold Dynamic Distribution Adaptation}

In this section, we introduce the learning of DDA through manifold learning. We propose Manifold Dynamic Distribution Adaptation (MDDA) as shown in Fig.~\ref{fig-sub-manifold}. Manifold feature learning can serve as the feature learning step to mitigate the influence of feature distortion~\cite{baktashmotlagh2013unsupervised} in transfer learning. The features in manifold space can reflect a more detailed structure and property of the domains, thus avoiding feature distortion. MDDA learns $g(\cdot)$ in the \textit{Grassmann} manifold $\mathbb{G}(d)$~\cite{hamm2008grassmann} since features in the manifold have some geometrical structures~\cite{belkin2006manifold,hamm2008grassmann} that can avoid distortion in the original space. In addition, $\mathbb{G}(d)$ can facilitate classifier learning by treating the original $d$-dimensional subspace (i.e. feature vector) as its basic elements. Feature transformation and distribution alignment often have efficient numerical forms (i.e., they can be represented as matrix operations easily) and facilitate domain adaptation on $\mathbb{G}(d)$~\cite{hamm2008grassmann}. There are several approaches to transform the features into $\mathbb{G}$ \cite{gopalan2011domain,baktashmotlagh2014domain}. We embed Geodesic Flow Kernel (GFK)~\cite{gong2012geodesic} to learn $g(\cdot)$ for its computational efficiency.

When learning manifold features, MDDA tries to model the domains with $d$-dimensional subspaces and then embed them into $\mathbb{G}(d)$. Let $\mathbf{P}_s$ and $\mathbf{P}_t$ denote the PCA subspaces for the source and the target domain, respectively. $\mathbb{G}$ can thus be regarded as a collection of all $d$-dimensional subspaces. Each original subspace can be seen as a point in $\mathbb{G}$. Therefore, the geodesic flow $\{\Phi(t):0 \leq t \leq 1\}$ between two points can be used to establish a path between the two subspaces, where $t$ denotes the calculus variant between two domains. If we let $\mathbf{P}_s=\Phi(0)$ and $\mathbf{P}_t=\Phi(1)$, then finding a geodesic flow from $\Phi(0)$ to $\Phi(1)$ equals to transforming the original features into an infinite-dimensional feature space, which eventually eliminates the domain shift. This approach can be seen as an incremental way of `walking' from $\Phi(0)$ to $\Phi(1)$. Specifically, the new features can be represented as $\mathbf{z}=g(\mathbf{x}) = \Phi(t)^T \mathbf{x}$. From \cite{gong2012geodesic}, the inner product of transformed features $\mathbf{z}_i$ and $\mathbf{z}_j$ gives rise to a positive semidefinite geodesic flow kernel
\begin{equation}
\label{equ-gfk}
\langle\mathbf{z}_i,\mathbf{z}_j\rangle= \int_{0}^{1} (\Phi(t)^T \mathbf{x}_i)^T (\Phi(t)^T \mathbf{x}_j) \, dt = \mathbf{x}^T_i \mathbf{G} \mathbf{x}_j.
\end{equation}

The geodesic flow can be parameterized as
\begin{equation}
	\mathbf { \Phi } ( t ) = \mathbf { P } _ { s } \mathbf { U } _ { 1 } \mathbf { \Gamma } ( t ) - \mathbf { R } _ { s } \mathbf { U } _ { 2 } \mathbf { \Sigma } ( t ) = \left[ \begin{array} { l l } { \mathbf { P } _ { s } } & { \mathbf { R } _ { s } } \end{array} \right] \left[ \begin{array} { c c } { \mathbf { U } _ { 1 } } & { 0 } \\ { 0 } & { \mathbf { U } _ { 2 } } \end{array} \right] \left[ \begin{array} { c } { \mathbf { \Gamma } ( t ) } \\ { \mathbf { -\Sigma } ( t ) } \end{array} \right],
\end{equation}
where $\mathbf{R}_s \in \mathbb{R}^{D \times d}$ presents the orthogonal complements to $\mathbf{P}_s$. $\mathbf{U}_1 \in \mathbb{R}^{D \times d}$ and $\mathbf{U}_2 \in \mathbb{R}^{D \times d}$ are two orthonormal matrices that can be computed by singular value decomposition (SVD)
\begin{equation}
	\mathbf { P } _ { S } ^ { \mathrm { T } } \mathbf { P } _ { T } = \mathbf { U } _ { 1 } \mathbf { \Gamma } \mathbf { V } ^ { \mathrm { T } } , \mathbf { R } _ { S } ^ { \mathrm { T } } \mathbf { P } _ { T } = - \mathbf { U } _ { 2 } \boldsymbol { \Sigma } \mathbf { V } ^ { \mathbf { T } }
\end{equation}

According to GFK~\cite{gong2012geodesic}, the geodesic flow kernel $\mathbf{G}$ can be calculated by
\begin{equation}
	\mathbf { G } = \left[ \begin{array} { l l l } { \mathbf { P } _ { s } \mathbf { U } _ { 1 } } & { \mathbf { R } _ { s } \mathbf { U } _ { 2 } } \end{array} \right] \left[ \begin{array} { l l } { \mathbf { \Lambda } _ { 1 } } & { \mathbf { \Lambda } _ { 2 } } \\ { \mathbf { \Lambda } _ { 2 } } & { \mathbf { \Lambda } _ { 3 } } \end{array} \right] \left[ \begin{array} { c } { \mathbf { U } _ { 1 } ^ { \mathrm { T } } \mathbf { P } _ { s } ^ { \mathrm { T } } } \\ { \mathbf { U } _ { 2 } ^ { \top } \mathbf { R } _ { s } ^ { \mathrm { T } } } \end{array} \right],
\end{equation}
where $\mathbf { \Lambda }_1, \mathbf { \Lambda }_2, \mathbf { \Lambda }_3$ are three diagonal matrices with elements
\begin{equation}
	\lambda _ { 1 i } = 1 + \frac { \sin \left( 2 \theta _ { i } \right) } { 2 \theta _ { i } } , \lambda _ { 2 i } = \frac { \cos \left( 2 \theta _ { i } \right) - 1 } { 2 \theta _ { i } } , \lambda _ { 3 i } = 1 - \frac { \sin \left( 2 \theta _ { i } \right) } { 2 \theta _ { i } }.
\end{equation}

Thus, the features in the original space can be transformed into Grassmann manifold with $\mathbf{z}=g(\mathbf{x}) = \sqrt{\mathbf{G}}\mathbf{x}$.

After manifold feature learning and dynamic distribution alignment, $f$ can be learned by summarizing SRM over $\Omega_s$ and distribution alignment. By adopting the square loss $l_2$, $f$ can be expressed as
\begin{equation}
\label{equ-f}
f = \mathop{\arg\min}_{f \in \mathcal{H}_{K}} \sum_{i=1}^{n} (y_i - f(\mathbf{z}_i))^2 + \eta ||f||^2_K + \lambda \overline{D_f}(\Omega_s,\Omega_t) + \rho R_f(\Omega_s,\Omega_t).
\end{equation}

In order to perform efficient learning, we now further reformulate each term.

\textbf{SRM on the Source Domain:} 
Using the representer theorem~\cite{belkin2006manifold}, $f$ can be expanded as 
\begin{equation}
\label{equ-repr}
f(\mathbf{z})=\sum_{i=1}^{n+m} \beta_i K(\mathbf{z}_i,\mathbf{z}),
\end{equation}
where $\bm{\beta}=(\beta_1,\beta_2,\cdots)^T \in \mathbb{R}^{(n + m) \times 1}$ is the coefficients vector and $K$ is a kernel. Then, SRM on $\Omega_s$ becomes
\begin{equation}
\label{equ-risk2}
\sum_{i=1}^{n} (y_i - f(\mathbf{z}_i))^2 + \eta ||f||^2_K 
= \sum_{i=1}^{n+m} \mathbf{A}_{ii}(y_i - f(\mathbf{z}_i))^2 + \eta ||f||^2_K
= ||(\mathbf{Y} - \bm{\beta}^T \mathbf{K}) \mathbf{A}||^2_{F} + \eta \mathrm{tr}(\bm{\beta}^T \mathbf{K} \bm{\beta}),
\end{equation}
where $||\cdot||_F$ is the Frobenious norm. $\mathbf{K} \in \mathbb{R}^{(n+m) \times (n+m)}$ is the kernel matrix with $\mathbf{K}_{ij}=K(\mathbf{z}_i,\mathbf{z}_j)$, and $\mathbf{A} \in \mathbb{R}^{(n+m) \times (n+m)}$ is a diagonal domain indicator matrix with $\mathbf{A}_{ii}=1$ if $i \in \Omega_s$, otherwise $\mathbf{A}_{ii}=0$. $\mathbf{Y}=[y_1,\cdots,y_{n+m}]$ is the label matrix from the source and the target domains. $\mathrm{tr(\cdot)}$ denotes the trace operation. $\eta$ is the shrinkage parameter. Although the labels for $\Omega_t$ are unavailable, they can be filtered out by the indicator matrix $\mathbf{A}$.

\textbf{Dynamic distribution adaptation: }
Using the representer theorem and kernel tricks, dynamic distribution alignment in equation~(\ref{equ-da1}) becomes
\begin{equation}
\label{equ-da}
\overline{D_{f}}(\Omega_s,\Omega_t)=\mathrm{tr} \left(\bm{\beta}^T \mathbf{K} \mathbf{M} \mathbf{K} \bm{\beta} \right)
\end{equation}
where $\mathbf{M}=(1-\mu)\mathbf{M}_0 + \mu \sum_{c=1}^{C} \mathbf{M}_c$ is the MMD matrix with its element calculated by
\begin{equation}
\label{equ-mo}
(\mathbf{M}_0)_{ij}=\begin{cases}
\frac{1}{n^2},  & \mathbf{z}_i,\mathbf{z}_j \in \Omega_s\\ 
\frac{1}{m^2}, & \mathbf{z}_i,\mathbf{z}_j \in \Omega_t\\ 
-\frac{1}{mn}, & \text{otherwise} 
\end{cases}
\end{equation}
\begin{equation}
\label{equ-mc}
(\mathbf{M}_c)_{ij}=\begin{cases}
\frac{1}{n^2_c},  & \mathbf{z}_i,\mathbf{z}_j \in \mathcal{D}^{(c)}_s\\ 
\frac{1}{m^2_c}, & \mathbf{z}_i,\mathbf{z}_j \in \mathcal{D}^{(c)}_t\\ 
-\frac{1}{m_c n_c}, & \begin{cases}
\mathbf{z}_i \in \mathcal{D}^{(c)}_s ,\mathbf{z}_j \in \mathcal{D}^{(c)}_t \\ 
\mathbf{z}_i \in \mathcal{D}^{(c)}_t ,\mathbf{z}_j \in \mathcal{D}^{(c)}_s
\end{cases}\\
0, & \text{otherwise}
\end{cases}
\end{equation}
where $n_c=|\mathcal{D}^{(c)}_s|$ and $m_c=|\mathcal{D}^{(c)}_t|$.

\textbf{Laplacian Regularization: } 
Additionally, we add a Laplacian regularization term to further exploit the similar geometrical property of nearest points in manifold $\mathbb{G}$~\cite{belkin2006manifold}. We denote the pair-wise affinity matrix as
\begin{equation}
\mathbf{W}_{ij} = \begin{cases}
\mathrm{sim}(\mathbf{z}_i,\mathbf{z}_j), & \mathbf{z}_i \in \mathcal{N}_p(\mathbf{z}_j) \text{ or } \mathbf{z}_j \in \mathcal{N}_p(\mathbf{z}_i) \\
0, & \text{otherwise}
\end{cases},
\end{equation}
where $\mathrm{sim}(\cdot,\cdot)$ is a similarity function~(such as cosine distance) for measuring the distance between two points. $\mathcal{N}_p(\mathbf{z}_i)$ denotes the set of $p$-nearest neighbors to point $\mathbf{z}_i$. $p$ is a free parameter and must be set in the method. By introducing Laplacian matrix $\mathbf{L}=\mathbf{D} - \mathbf{W}$ with diagonal matrix $\mathbf{D}_{ii}=\sum_{j=1}^{n+m} \mathbf{W}_{ij}$, the final regularization can be expressed as
\begin{equation}
\label{equ-lap}
R_f(\Omega_s,\Omega_t)=\sum_{i,j=1}^{n+m} \mathbf{W}_{ij} (f(\mathbf{z}_i)-f(\mathbf{z}_j))^2
=\sum_{i,j=1}^{n+m} f(\mathbf{z}_i) \mathbf{L}_{ij} f(\mathbf{z}_j)
=\mathrm{tr} \left(\bm{\beta}^T \mathbf{K} \mathbf{L} \mathbf{K} \bm{\beta}\right).
\end{equation}

\textbf{Overall Reformulation:} By combining equations~(\ref{equ-risk2}), (\ref{equ-da}) and (\ref{equ-lap}), $f$ in equation (\ref{equ-f}) can be reformulated as
\begin{equation}
\label{equ-final}
f=\mathop{\arg\min}_{f \in \mathcal{H}_{K}}||(\mathbf{Y} - \bm{\beta}^T \mathbf{K}) \mathbf{A}||^2_{F} + \eta \, \mathrm{tr}(\bm{\beta}^T \mathbf{K} \bm{\beta}) + \mathrm{tr}\left(\bm{\beta}^T \mathbf{K}(\lambda \mathbf{M} + \rho \mathbf{L}) \mathbf{K} \bm{\beta} \right).
\end{equation}

Setting derivative $\partial f/ \partial \bm{\beta}=0$, we obtain the solution 
\begin{equation}
\label{equ-solution}
\bm{\beta}^\star= ((\mathbf{A} + \lambda \mathbf{M} + \rho \mathbf{L}) \mathbf{K} + \eta \mathbf{I})^{-1} \mathbf{A} \mathbf{Y}^T.
\end{equation}

\begin{algorithm}[t!] 
	\caption{MDDA: Manifold Dynamic Distribution Adaptation}  
	\label{algo-mdda}  
	\renewcommand{\algorithmicrequire}{\textbf{Input:}} 
	\renewcommand{\algorithmicensure}{\textbf{Output:}}
	\begin{algorithmic}[1]  
		\REQUIRE 
		Data matrix $\mathbf{X}=[\mathbf{X}_s,\mathbf{X}_t]$, source domain labels $\mathbf{y}_s$, manifold subspace dimension $d$, regularization parameters $\lambda,\eta,\rho$, and \#neighbor $p$.\\
		\ENSURE 
		Classifier $f$.\\
		\STATE Learn manifold feature transformation kernel~$\mathbf{G}$ via equation~(\ref{equ-gfk}), and get manifold feature $\mathbf{Z}=\sqrt{\mathbf{G}} \mathbf{X}$.
		\STATE Train a base classifier using $\Omega_s$, then apply prediction on $\Omega_t$ to get its soft labels $\hat{y}_t$. 
		\STATE Construct kernel $\mathbf{K}$ using transformed features $\mathbf{Z}_s=\mathbf{Z}_{1:n,:}$ and $\mathbf{Z}_t=\mathbf{Z}_{n+1:n+m,:}$.
		\REPEAT 
		\STATE Calculate the adaptive factor $\hat{\mu}$ using equation~(\ref{eq-mu}).
		and compute $\mathbf{M}_0$ and $\mathbf{M}_c$ by Eq.~(\ref{equ-mo}) and (\ref{equ-mc}).
		\STATE Compute $\bm{\beta}^\star$ by solving equation~(\ref{equ-solution}) and obtain $f$ via the representer theorem in Eq.~(\ref{equ-repr}).
		\STATE Update the soft labels of $\Omega_t$: $\hat{y}_t=f(\mathbf{Z}_t)$.
		\UNTIL{Convergence}  
		\RETURN Classifier $f$.  
	\end{algorithmic}  
\end{algorithm}

\label{sec-deep}
\subsection{DDAN: Dynamic Distribution Adaptation Network}

In this section, we propose Dynamic Distribution Adaptation Network (DDAN) to perform end-to-end learning of not only the feature learning function $g(\cdot)$, but the classifier $f$. DDAN is able to leverage the ability of the recent advance of deep neural networks in learning representative features through end-to-end training~\cite{bengio2013representation}. Specifically, we exploit a backbone network to learn useful feature representations, while simultaneously performing domain adaptation using DDA.

The network architecture of DDAN is shown in Fig~\ref{fig-sub-deep}. Firstly, the samples from the source and target domains serve as the inputs into the deep neural networks. Secondly, the CNN network (the purple part) such as AlexNet~\cite{krizhevsky2012imagenet} and ResNet~\cite{he2016deep} can extract high-level features from the inputs. Thirdly, the features are going through a fully-connected layer (the blue part) to perform soft-max classification to obtain the labels $y$. The novel contribution here is to align the source and target domain features using the dynamic distribution alignment (DDA, the yellow part).

Adopting the DDA, the learning objective of DDAN can be expressed as

\begin{equation}
\label{eq-deepMDDA}
f = \min_{\bm{\Theta}} \sum_{i=1}^{n} J(f(\mathbf{x}^s_i),y^s_i) + \lambda \overline{D_{f}}(\Omega_s,\Omega_t) + \rho R_f(\Omega_s,\Omega_t),
\end{equation}
where $J(\cdot,\cdot)$ is the cross-entropy loss function and $\bm{\Theta}=\{\mathbf{w},b\}$ containing the weight and bias parameters of the neural network. Note that DDAN is based on the deep neural network, so instead of using the whole domain data, we use the batch data by following the mini-batch stochastic gradient descent (SGD) training procedure. Therefore, the dynamic distribution adaptation is only calculated between batches rather than whole domains. This is more practical and efficient in real applications where the data are coming in a streaming manner.


Most MMD based deep transfer learning methods~\cite{tzeng2014deep} are based on Eq.~(\ref{eq-mmd}) and only adopted the linear kernel for simplicity. Since the formulation in Eq.~(\ref{eq-mmd}) is based on pairwise similarity and is computed in quadratic time complexity, it is prohibitively time-consuming for using mini-batch stochastic gradient decent (SGD) in CNN-based transfer learning methods. Gretton \textit{et al.}~\cite{gretton2012optimal} further suggest an unbiased approximation of MMD with linear complexity. Without loss of generality, by assuming $M = N$, MMD can then be computed as
\begin{equation}
	\label{eq-linear-mmd}
	MMD^2_l(s,t) = \frac{2}{M} \sum_{i=1}^{M / 2} h_l(\mathbf{z}_i),
\end{equation}
where $h_l$ is an operator defined on a quad-tuple $\mathbf{z}_i = \left( \mathbf { x } _ { 2 i - 1 } ^ { s } , \mathbf { x } _ { 2 i } ^ { s } , \mathbf { x } _ { 2 j - 1 } ^ { t } , \mathbf { x } _ { 2 j } ^ { t } \right)$,
\begin{equation}
	h _ { l } \left( \mathbf { z } _ { i } \right) = k \left( \mathbf { x } _ { 2 i - 1 } ^ { s } , \mathbf { x } _ { 2 i } ^ { s } \right) + k \left( \mathbf { x } _ { 2 j - 1 } ^ { t } , \mathbf { x } _ { 2 j } ^ { t } \right)  - k \left( \mathbf { x } _ { 2 i - 1 } ^ { s } , \mathbf { x } _ { 2 j } ^ { t } \right) - k \left( \mathbf { x } _ { 2 i } ^ { s } , \mathbf { x } _ { 2 j - 1 } ^ { t } \right)
\end{equation}

The approximation in Eq.~\ref{eq-linear-mmd} takes a summation form and is suitable for gradient computation in a mini-batch manner.

The gradient of the parameters can be computed as:

\begin{equation}
\label{eq-gradient}
\Delta_{\bm{\Theta}} = \frac{\partial J(\cdot,\cdot)}{\partial \bm{\Theta}} + \lambda \frac{\partial \overline{D_{f}}(\cdot,\cdot)}{\partial \bm{\Theta}} + \rho \frac{\partial R_{f}(\cdot,\cdot)}{\partial \bm{\Theta}}.
\end{equation}

\textbf{Updating $\bm{\mu}$:} Another important aspect is to dynamically update $\mu$ in DDAN. Similar to the above mini-batch learning of MMD distance, it seems natural to calculate $\mu$ after each mini-batch learning. However, the labels on the source domain and the pseudo labels on the target domain are likely to be inconsistent after mini-batch learning. For instance, assume that the batch size for both domains is $b_{size} = 32$ and the total class number $|\mathcal{C}| = 31$. Then, after a forward operation for one batch, we can obtain 32 pseudo labels for the target domain. Since $b_{size}$ is rather close to $|\mathcal{C}|$, it is highly likely that the mini-batch labels for both domains do not match, which will easily lead to the mode collapse or gradient exploding problem.

In order to avoid this problem, we propose to update $\mu$ after each epoch of iteration, rather than each mini-batch data. In fact, this step is very similar to that of MDDA, which uses all the data to perform learning. The learning process of DDAN is summarized in Algorithm~\ref{algo-DDAN}.

\begin{algorithm}[t!] 
	\caption{DDAN: Dynamic Distribution Adaptation Network}  
	\label{algo-DDAN}  
	\renewcommand{\algorithmicrequire}{\textbf{Input:}} 
	\renewcommand{\algorithmicensure}{\textbf{Output:}}
	\begin{algorithmic}[1]  
		\REQUIRE 
		Source domain $(\mathbf{x}^s,\mathbf{y}^s)$, target domain data $\mathbf{x}^t$, regularization parameters $\lambda,\rho$, and \#neighbor $p$.\\
		\ENSURE 
		Classifier $f$.\\
		\REPEAT
		\STATE Sample a mini-batch data from both the source and target domain
		\STATE Feed the mini-batch data into the network and get the pseudo labels for $\Omega_t$
		\STATE Update the parameters $\{\Theta,\bm{b}\}$ by computing the mini-batch gradient according to Eq.~(\ref{eq-gradient}).
		\STATE After an epoch, calculate $\mu$ using Eq.~(\ref{eq-mu}) and calculate the loss
		\UNTIL{Convergence}  
		\RETURN Classifier $f$.  
	\end{algorithmic}  
\end{algorithm}

\subsection{Discussions}

Both MDDA and DDAN are generic learning methods that are suitable for all transfer classification problems. In this section, we briefly discuss their differences.

MDDA is for traditional learning, while DDAN is for deep learning. Compared to DDAN which is designed for deep neural networks, MDDA can be easily applied on the small resource constraint devices. One possible limitation of MDDA maybe that it relies on certain feature extraction methods. For instance, on image dataset, we probably need to extract SIFT, SURF, or HOG features. Luckily, MDDA can also use the deep features extracted by deep neural networks such as AlexNet~\cite{krizhevsky2012imagenet} and ResNet~\cite{he2016deep}. MDDA has a very useful property: it can learn the cross-domain function directly without the need for explicit classifier training. This makes it significantly more advantageous compared to most existing work such as JGSA~\cite{zhang2017joint} and SCA~\cite{ghifary2017scatter} which needs to train an extra classifier after learning transferrable features.

On the other hand, DDAN is suitable for cloud computing. The model of DDAN can be trained in an end-to-end manner and then be used for inference on the device. DDAN does not need extra feature extraction and classifier training procedure. All steps can be unified in one single deep neural network. This advantage makes it useful for large-scale datasets, which will probably result in prohibitive computations for MDDA.

\section{Experiments and Evaluations}
\label{sec-exp}
In this section, we evaluate the performance of MDDA through extensive experiments on large-scale public datasets. The source code for MDDA is available at \url{http://transferlearning.xyz}. We will focus on evaluating the performance of MDDA since most of our contributions can be covered by MDDA. In the last part of this section, we will evaluate the performance of the deep version of MDDA.

\subsection{Experimental Setup}

\subsubsection{Datasets}
\label{sec-exp-data}

\begin{figure}[htbp]
	\centering
	\includegraphics[scale=.4]{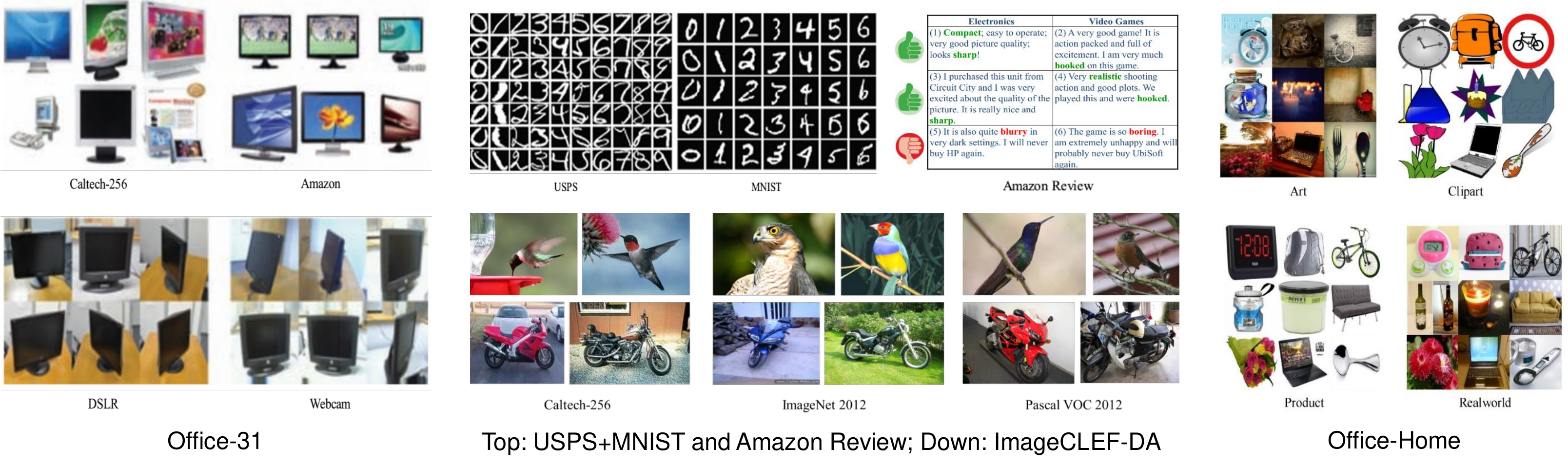}
	\vspace{-.1in}
	\caption{Samples from the datasets in this paper}
	\label{fig-dataset}
	\vspace{-.2in}
\end{figure}

We adopted five public image datasets: USPS+MNIST, Amazon review~\cite{blitzer2006domain}, Office-31~\cite{saenko2010adapting}, ImageCLEF-DA~\cite{long2017deep}, and Office-Home~\cite{venkateswara2017deep}. These datasets are popular for benchmarking domain adaptation algorithms and have been widely adopted in most existing work such as~\cite{zhang2017joint,long2018conditional,zhang2018collaborative,chen2018joint}. Fig.~\ref{fig-dataset} shows some samples of the datasets, and Table~\ref{tb-dataset} lists their statistics.

\begin{table}[htbp]
	\centering
	\caption{Statistics of the five benchmark datasets.}
	\vspace{-.1in}
	\label{tb-dataset}
	\resizebox{.9\textwidth}{!}{
	\begin{tabular}{|c|c|c|c|c|c|}
		\hline
		\textbf{Dataset} & \textbf{\#Sample} & \textbf{\#Feature for MDDA} & \textbf{\#Class} & \textbf{Domain} & \textbf{Type} \\ \hline \hline
		USPS+MNIST & 3,800 & 256 & 10 & U, M  & Digit\\ \hline
		Amazon review & 1,123 & 400 & 2 & B, D, E, K  & Text\\ \hline
		Office-31 & 4,110 & 2,048 & 31 & A, W, D  & Image\\ \hline	
		ImageCLEF DA & 1,800 & 2,048 & 12 & C, I, P  & Image \\ \hline
		Office-Home & 15,500 & 2,048 & 65 & Ar, Cl, Pr, Rw  & Image \\ \hline
	\end{tabular}
}
\end{table}

\textbf{USPS} (U) and \textbf{MNIST} (M) are standard digit recognition datasets containing handwritten digits from 0-9. Since the same digits across two datasets follow different distributions, it is necessary to perform domain adaptation. USPS consists of 7,291 training images and 2,007 test images of size 16 $\times$ 16. MNIST consists of 60,000 training images and 10,000 test images of size 28 $\times$ 28. We construct two tasks: U $\rightarrow$ M and M $\rightarrow$ U. In the rest of the paper, we use $A \rightarrow B$ to denote the knowledge transfer from source domain \textit{A} to the target domain \textit{B}.

\textbf{Amazon review}~\cite{blitzer2006domain} is the benchmark dataset for cross-domain sentiment analysis. This dataset includes reviews about the Kitchen appliances~(K), DVDs~(D), Books~(B) and Electronics~(E). The reviews of each product can be regarded as data from the same domain. There are 1000 positive and 1000 negative instances on each domain. Transfer learning can be conducted between any two domains, leading to 12 tasks.

\textbf{Office-31}~\cite{saenko2010adapting} consists of three real-world object domains: \textbf{Amazon} (A), \textbf{Webcam} (W) and \textbf{DSLR} (D). It has 4,652 images with 31 categories. \textbf{Caltech-256} (C) contains 30,607 images and 256 categories. We constructed 6 tasks: A $\rightarrow$ D, A $\rightarrow$ W, D $\rightarrow$ A, D $\rightarrow$ W, W $\rightarrow$ A, W $\rightarrow$ D.

\textbf{ImageCLEF-DA}~\cite{long2017deep} is a dataset presented in the ImageCLEF 2014 domain adaptation challenge. It is composed by selecting the 12 common classes shared by three public
datasets (domains): Caltech-256 (C), ImageNet ILSVRC 2012 (I), and Pascal VOC 2012 (P). There
are 50 images in each category and 600 images in each domain, while Office-31 has different domain
sizes. We permute domains and build 6 transfer tasks: C $\rightarrow$ I, C $\rightarrow$ P, I $\rightarrow$ C, I $\rightarrow$ P, P $\rightarrow$ C,  P $\rightarrow$ I.

\textbf{Office-Home}~\cite{venkateswara2017deep} is a new dataset which consists of 15,588 images from 4 different domains: Artistic images (Ar), Clip Art (Cl), Product images (Pr), and Real-World images (Rw). For each domain, the dataset contains images of 65 object categories collected from office and home settings. We use all domains and construct 12 transfer learning tasks: Ar $\rightarrow$ Cl, $\cdots$, Rw $\rightarrow$ Pr.

In total, we constructed $2 + 12 + 6 + 6 + 12 = 38$ tasks.

\subsubsection{State-of-the-art Comparison Methods}
We compared the performance of MDDA with several state-of-the-art traditional and deep transfer learning approaches.

Traditional transfer learning methods:

\begin{itemize}[noitemsep]
	\item \textbf{NN}, \textbf{SVM}, and \textbf{PCA}
	\item \textbf{TCA}: Transfer Component Analysis~\cite{pan2011domain}, which performs marginal distribution alignment
	\item \textbf{GFK}: Geodesic Flow Kernel~\cite{gong2012geodesic}, which performs manifold feature learning
	\item \textbf{JDA}: Joint Distribution Adaptation~\cite{long2013transfer}, which adapts marginal \& conditional distribution
	\item \textbf{CORAL}: CORrelation Alignment~\cite{sun2016return}, which performs second-order subspace alignment
	\item \textbf{SCA}: Scatter Component Analysis~\cite{ghifary2017scatter}, which adapts scatters in subspace
	\item \textbf{JGSA}: Joint Geometrical and Statistical Alignment~\cite{zhang2017joint}, which aligns marginal \& conditional distributions with label propagation
\end{itemize}

Deep transfer learning methods:

\begin{itemize}[noitemsep]
	\item \textbf{AlexNet}~\cite{krizhevsky2012imagenet} and \textbf{ResNet}~\cite{he2016deep}, as baseline networks
	\item \textbf{DDC}: Deep Domain Confusion~\cite{tzeng2014deep}, which is a single-layer deep adaptation method
	\item \textbf{DAN}: Deep Adaptation Network~\cite{long2015learning}, which is a multi-layer adaptation method
	\item \textbf{DANN}: Domain Adversarial Neural Network~\cite{ganin2014unsupervised}, which is a adversarial deep neural network
	\item \textbf{ADDA}: Adversarial Discriminative Domain Adaptation~\cite{tzeng2017adversarial}, which is a general framework for adversarial transfer learning
	\item \textbf{JAN}: Joint Adaptation Networks~\cite{long2017deep}, which is a deep network with joint MMD distance
	\item \textbf{CAN}: Collaborative and Adversarial Network~\cite{zhang2018collaborative}, which is based on joint training
	\item \textbf{CDAN}: Conditional Domain Adversarial Networks~\cite{long2018conditional}, which is a conditional network
\end{itemize}

\subsubsection{Implementation Details}

MDDA requires to extract features from the raw inputs. For USPS+MNIST datasets, we adopted the 256 SURF features by following existing work~\cite{zhang2017joint,wang2017balanced}. For Amazon review dataset, we follow the feature generation method to exploit marginalized denoising autoencoders~\cite{chen2012marginalized} to improve the feature representations. For Office-31, ImageCLEF DA, and Office-Home datasets, we adopted the 2048 fine-tuned ResNet-50 features for a fair comparison. As for DDAN, we only report its results on three image datasets since USPS+MNIST and Amazon review datasets are rather simple to transfer. DDAN is able to take the original image data as inputs. We also adopted ResNet-50 as the baseline network for fair comparison~\cite{zhang2018collaborative,long2018conditional}.

For the comparison methods, we either cite the results reported in their original papers or run experiments using their publicly available codes. As for MDDA, we set the manifold feature dimension $d=30,30,50,60,200$ for the five datasets, respectively. The number of iteration is set to $T=10$. We use the RBF kernel with the bandwidth set to be the variance of inputs. The regularization parameters are set as $p=10,\lambda=4.5,\eta=0.1$, and $\rho=1$. Additionally, the experiments on parameter sensitivity and convergence analysis in Section~\ref{sec-para} indicate that the performance of MDDA and DDAN stays robust with a wide range of parameter choices. For DDAN, we set the learning rate to be 0.01 with the batch size to be 32 and a weight decay of $5e-4$. Other parameters are tuned by following transfer cross validation~\cite{zhong2009cross}.

Although MDDA is easy to use, and its parameters do not have to be fine-tuned. For research purpose, we also investigate how to further tune those parameters. We choose parameters according to the following rules. Firstly, SRM on source domain is very important. Thus, we prefer a small $\eta$ to make sure MDDA does not degenerate. Secondly, distribution alignment is required by SRM. Thus, we choose a slightly larger $\lambda$ to make it effective. Thirdly, we choose $\rho$ by following the existing work~\cite{belkin2006manifold}. Fourthly, $p$ is set following~\cite{cai2011graph}.

We adopt classification accuracy on $\Omega_t$ as the evaluation metric, which is widely used in existing literature \cite{pan2011domain,wang2017balanced,gong2012geodesic}:
\begin{equation}
	Accuracy = \frac{\left|\mathbf{x} : \mathbf{x} \in \Omega_{t} \wedge \hat{y}(\mathbf{x})=y(\mathbf{x})\right|}{\left|\mathbf{x} : \mathbf{x} \in \Omega_{t}\right|}.
\end{equation}

\subsection{Results and Analysis}

\subsubsection{Results on Digit Datasets}
The classification results on USPS+MNIST datasets are shown in Table~\ref{tb-usps}. On the digit recognition tasks, MDDA outperforms the best method JGSA by a large margin of $\bm{8.9\%}$. These results clearly indicate that MDDA significantly outperforms existing methods.

Moreover, the performances of distribution alignment methods (TCA, JDA, and JGSA) and subspace learning methods~(GFK, CORAL, and SCA) were generally inferior to MDDA. Each method has its limitations and cannot handle domain adaptation in specific tasks, especially with degenerated feature transformation and unevaluated distribution alignment. After manifold or subsapce learning, there still exists large domain shift~\cite{baktashmotlagh2013unsupervised}; while feature distortion will undermine the distribution alignment methods. 

\begin{table}[htbp]
	\centering
	\caption{Classification accuracy (\%) on USPS-MNIST datasets with SURF features}
	\vspace{-.1in}
	\label{tb-usps}
	\resizebox{.7\textwidth}{!}{
		\begin{tabular}{|c|c|c|c|c|c|c|c|c|c|}
			\hline
			Task & 1NN & SVM & TCA & GFK & JDA & CORAL & SCA & JGSA & MDDA \\ \hline \hline
			U $\rightarrow$ M & 44.7 & 62.2 & 51.2 & 46.5 & 59.7 & 30.5 & 48.0 & 68.2 & \textbf{76.8} \\ \hline
			M $\rightarrow$ U & 65.9 & 68.2 & 56.3 & 61.2 & 67.3 & 49.2 & 65.1 & 80.4 & \textbf{89.6} \\ \hline \hline
			AVG & 55.3 &  65.2 &  53.8 &  53.9 &  63.5 &  39.9 &  56.6 &  74.3 &  \textbf{83.2} \\ \hline 
		\end{tabular}
	}
	\vspace{-.1in}
\end{table}

\subsubsection{Results on Sentiment Analysis Dataset}

The results on Amazon review datasets are shown in Table~\ref{tb-amazon}. From the results, we can observe that our proposed MDDA outperforms the best baseline method CORAL by a large margin of \textbf{6.0}\%. This clearly indicated that MDDA is able to dramatically reduce the divergence between different text domains.

\begin{table}[htbp]
	\caption{Classification accuracy (\%) on Amazon review dataset}
	\label{tb-amazon}
	\vspace{-.1in}
	\resizebox{.65\textwidth}{!}{
	\begin{tabular}{|c|c|c|c|c|c|c|c|c|}
		\hline
		Method & 1NN & TCA & GFK & SA & JDA & CORAL & JGSA & MDDA \\ \hline \hline
		B $\rightarrow$ D & 49.6 & 63.6 & 66.4 & 67.0 & 64.2 & 71.6 & 66.6 & \textbf{77.8} \\ \hline
		B $\rightarrow$ E & 49.8 & 60.9 & 65.5 & 70.8 & 62.1 & 65.1 & 75.0 & \textbf{80.0} \\ \hline
		B $\rightarrow$ K & 50.3 & 64.2 & 69.2 & 72.2 & 65.4 & 67.3 & 72.1 & \textbf{79.9} \\ \hline
		D $\rightarrow$ B & 53.3 & 63.3 & 66.3 & 67.5 & 62.4 & 70.1 & 55.5 & \textbf{74.7} \\ \hline
		D $\rightarrow$ E & 51.0 & 64.2 & 63.7 & 67.1 & 66.3 & 65.6 & 67.3 & \textbf{80.4} \\ \hline
		D $\rightarrow$ K & 53.1 & 69.1 & 67.7 & 69.4 & 68.9 & 67.1 & 65.6 & \textbf{81.0} \\ \hline
		E $\rightarrow$ B & 50.8 & 59.5 & 62.4 & 61.4 & 59.2 & 67.1 & 51.6 & \textbf{63.8} \\ \hline
		E $\rightarrow$ D & 50.9 & 62.1 & 63.4 & 64.9 & 61.6 & 66.2 & 50.8 & \textbf{62.5} \\ \hline
		E $\rightarrow$ K & 51.2 & 74.8 & 73.8 & 70.4 & 74.7 & 77.6 & 55.0 & \textbf{84.4} \\ \hline
		K $\rightarrow$ B & 52.2 & 64.1 & 65.5 & 64.4 & 62.7 & 68.2 & 58.3 & \textbf{63.5} \\ \hline
		K $\rightarrow$ D & 51.2 & 65.4 & 65.0 & 64.6 & 64.3 & 68.9 & 56.4 & \textbf{72.2} \\ \hline
		K $\rightarrow$ E & 52.3 & 74.5 & 73.0 & 68.2 & 74.0 & 75.4 & 51.7 & \textbf{80.7} \\ \hline \hline
		AVG & 51.3 & 65.5 & 66.8 & 67.3 & 65.5 & 69.1 & 60.5 & \textbf{75.1} \\ \hline
	\end{tabular}
}
\vspace{-.1in}
\end{table}

\subsubsection{Results on Image Datasets}

The classification accuracy results on the Office-31, ImageCLEF DA, and Office-Home datasets are shown in Tables~\ref{tb-office31-resnet}, \ref{tb-imageclef}, and \ref{tb-officehome}, respectively. From those results, we can make the following observations.

Firstly, MDDA outperforms all other traditional and deep comparison methods in most tasks (20/24 tasks). The average classification accuracy achieved by MDDA on all the image tasks is $\bm{77.3}$\textbf{\%}. Specifically, on the hardest Office-Home dataset, MDDA significantly outperforms the latest deep transfer learning method CDAN~\cite{long2018conditional} by $\bm{4.5\%}$, which clearly demonstrates the effectiveness of MDDA. The results indicates that MDDA is capable of significantly reducing the distribution divergence in domain adaptation problems. 

Secondly, DDAN also substantially outperforms all the traditional and deep methods on most tasks. Note that DDAN is only based on deep neural network without adversarial training, while other deep methods such as CAN~\cite{zhang2018collaborative} and CDAN~\cite{long2018conditional} all require to train an adversarial neural network, which clearly needs more time to converge. In this way, DDAN is much more efficient than these networks.

Thirdly, we also note that traditional methods such as TCA, JDA, and CORAL can also achieve good performance compared to ResNet, 1NN, and SVM. This clearly indicates the necessity of transfer learning when building models from two domains. Again, our proposed MDDA and DDAN can achieve the best performances.

\begin{table}[htbp]
	\centering
	\caption{Classification accuracy (\%) on Office-31 dataset with ResNet-50 as the baseline}
	\label{tb-office31-resnet}
	\resizebox{1\textwidth}{!}{
	\begin{tabular}{|c|c|c|c|c|c|c|c|c|c|c|c|c|c|c|c|}
		\hline
		Method & \multicolumn{3}{c|}{Baseline} & \multicolumn{4}{c|}{Traditional transfer learning} & \multicolumn{6}{c|}{Deep transfer learning} & \multicolumn{2}{c|}{DDA} \\ \hline
		Task & ResNet & 1NN & SVM & TCA & GFK & JDA & CORAL & DDC & DAN & DANN & ADDA & JAN & CAN & MDDA & DDAN \\ \hline \hline
		A $\rightarrow$ D & 68.9 & 79.1 & 76.9 & 74.1 & 77.9 & 80.7 & 81.5 & 76.5 & 78.6 & 79.7 & 77.8 & 84.7 & 85.5 & \textbf{86.3} & 84.9 \\ \hline
		A $\rightarrow$ W & 68.4 & 75.8 & 73.3 & 72.7 & 72.6 & 73.6 & 77.0 & 75.6 & 80.5 & 82.0 & 86.2 & 85.4 & 81.5 & 86.0 & \textbf{88.8} \\ \hline
		D $\rightarrow$ A & 62.5 & 60.2 & 64.1 & 61.7 & 62.3 & 64.7 & 65.9 & 62.2 & 63.6 & 68.2 & 69.5 & 68.6 & 65.9 & \textbf{72.1} & 65.3 \\ \hline
		D $\rightarrow$ W & 96.7 & 96.0 & 96.5 & 96.7 & 95.6 & 96.5 & 97.1 & 96.0 & 97.1 & 96.9 & 96.2 & 97.4 & \textbf{98.2} & 97.1 & 96.7 \\ \hline
		W $\rightarrow$ A & 60.7 & 59.9 & 64.9 & 60.9 & 62.8 & 63.1 & 64.3 & 61.5 & 62.8 & 67.4 & 68.9 & 70.0 & 63.4 & \textbf{73.2} & 65.0 \\ \hline
		W $\rightarrow$ D & 99.3 & 99.4 & 99.0 & 99.6 & 99.0 & 98.6 & 99.6 & 98.2 & 99.6 & 99.1 & 98.4 & 99.8 & 99.7 & 99.2 & \textbf{100} \\ \hline \hline
		AVG & 76.1 & 78.4 & 79.1 & 77.6 & 78.4 & 79.5 & 80.9 & 78.3 & 80.4 & 82.2 & 82.9 & 84.3 & 82.4 & \textbf{85.7} & 83.5 \\ \hline
	\end{tabular}
}
\end{table}

\begin{table}[htbp]
	\caption{Classification accuracy (\%) on ImageCLEF DA with ResNet-50 as baseline}
	\label{tb-imageclef}
	\resizebox{1\textwidth}{!}{
	\begin{tabular}{|c|c|c|c|c|c|c|c|c|c|c|c|c|c|c|}
		\hline
		Method & \multicolumn{3}{c|}{Baseline} & \multicolumn{4}{c|}{Traditional transfer learning} & \multicolumn{5}{c|}{Deep transfer learning} & \multicolumn{2}{c|}{DDA} \\ \hline
		Task & ResNet & 1NN & SVM & TCA & GFK & JDA & CORAL & DAN & DANN & JAN & CAN & CDAN & MDDA & DDAN \\ \hline \hline
		C $\rightarrow$ I & 78.0 & 83.5 & 86.0 & 89.3 & 86.3 & 90.8 & 83.0 & 86.3 & 87.0 & 89.5 & 89.5 & 91.2 & \textbf{92.0} & 91.0 \\ \hline
		C $\rightarrow$ P & 65.5 & 71.3 & 73.2 & 74.5 & 73.3 & 73.6 & 71.5 & 69.2 & 74.3 & 74.2 & 75.8 & 77.2 & \textbf{78.8} & 76.0 \\ \hline
		I $\rightarrow$ C & 91.5 & 89.0 & 91.2 & 93.2 & 93.0 & 94.0 & 88.7 & 92.8 & 96.2 & 94.7 & 94.2 & \textbf{96.7} & 95.7 & 94.0 \\ \hline
		I $\rightarrow$ P & 74.8 & 74.8 & 76.8 & 77.5 & 75.5 & 75.3 & 73.7 & 74.5 & 75.0 & 76.8 & 78.2 & 78.3 & \textbf{79.8} & 78.0 \\ \hline
		P $\rightarrow$ C & 91.2 & 76.2 & 85.8 & 83.7 & 82.3 & 83.5 & 72.0 & 89.8 & 91.5 & 91.7 & 89.2 & 93.7 & \textbf{95.5} & 92.7 \\ \hline
		P $\rightarrow$ I & 83.9 & 74.0 & 80.2 & 80.8 & 78.0 & 77.8 & 71.3 & 82.2 & 86.0 & 88.0 & 87.5 & 91.2 & \textbf{91.5} & 91.0 \\ \hline \hline
		AVG & 80.7 & 78.1 & 82.2 & 83.2 & 81.4 & 82.5 & 76.7 & 82.5 & 85.0 & 85.8 & 85.7 & 88.1 & \textbf{88.9} & 87.2 \\ \hline
	\end{tabular}
}
\end{table}

\begin{table}[htbp]
	\centering
	\caption{Classification accuracy (\%) on Office-Home dataset with ResNet-50 as baseline}
	\label{tb-officehome}
	\resizebox{1\textwidth}{!}{
	\begin{tabular}{|c|c|c|c|c|c|c|c|c|c|c|c|c|c|}
		\hline
		Method & \multicolumn{3}{c|}{Baseline} & \multicolumn{4}{c|}{Traditional transfer learning} & \multicolumn{4}{c|}{Deep transfer learning} & \multicolumn{2}{c|}{DDA} \\ \hline
		Task & ResNet & 1NN & SVM & TCA & GFK & JDA & CORAL & DAN & DANN & JAN & CDAN & MDDA & DDAN \\ \hline \hline
		Ar $\rightarrow$ Cl & 34.9 & 45.3 & 45.3 & 38.3 & 38.9 & 38.9 & 42.2 & 43.6 & 45.6 & 45.9 & 46.6 & \textbf{54.9} & 51.0 \\ \hline
		Ar $\rightarrow$ Pr & 50.0 & 60.1 & 65.4 & 58.7 & 57.1 & 54.8 & 59.1 & 57.0 & 59.3 & 61.2 & 65.9 & \textbf{75.9} & 66.0 \\ \hline
		Ar $\rightarrow$ Rw & 58.0 & 65.8 & 73.1 & 61.7 & 60.1 & 58.2 & 64.9 & 67.9 & 70.1 & 68.9 & 73.4 & \textbf{77.2} & 73.9 \\ \hline
		Cl $\rightarrow$ Ar & 37.4 & 45.7 & 43.6 & 39.3 & 38.7 & 36.2 & 46.4 & 45.8 & 47.0 & 50.4 & 55.7 & \textbf{58.1} & 57.0 \\ \hline
		Cl $\rightarrow$ Pr & 41.9 & 57.0 & 57.3 & 52.4 & 53.1 & 53.1 & 56.3 & 56.5 & 58.5 & 59.7 & 62.7 & \textbf{73.3} & 63.1 \\ \hline
		Cl $\rightarrow$ Rw & 46.2 & 58.7 & 60.2 & 56.0 & 55.5 & 50.2 & 58.3 & 60.4 & 60.9 & 61.0 & 64.2 & \textbf{71.5} & 65.1 \\ \hline
		Pr $\rightarrow$ Ar & 38.5 & 48.1 & 46.8 & 42.6 & 42.2 & 42.1 & 45.4 & 44.0 & 46.1 & 45.8 & 51.8 & \textbf{59.0} & 52.0 \\ \hline
		Pr $\rightarrow$ Cl & 31.2 & 42.9 & 39.1 & 37.5 & 37.6 & 38.2 & 41.2 & 43.6 & 43.7 & 43.4 & 49.1 & \textbf{52.6} & 48.4 \\ \hline
		Pr $\rightarrow$ Rw & 60.4 & 68.9 & 69.2 & 64.1 & 64.6 & 63.1 & 68.5 & 67.7 & 68.5 & 70.3 & 74.5 & \textbf{77.8} & 72.7 \\ \hline
		Rw $\rightarrow$ Ar & 53.9 & 60.8 & 61.1 & 52.6 & 53.8 & 50.2 & 60.1 & 63.1 & 63.2 & 63.9 & \textbf{68.2} & 67.9 & 65.1 \\ \hline
		Rw $\rightarrow$ Cl & 41.2 & 48.3 & 45.6 & 41.7 & 42.3 & 44.0 & 48.2 & 51.5 & 51.8 & 52.4 & 56.9 & \textbf{57.6} & 56.6 \\ \hline
		Rw $\rightarrow$ Pr & 59.9 & 74.7 & 75.9 & 70.5 & 70.6 & 68.2 & 73.1 & 74.3 & 76.8 & 76.8 & 80.7 & \textbf{81.8} & 78.9 \\ \hline \hline
		Avg & 46.1 & 56.4 & 56.9 & 51.3 & 51.2 & 49.8 & 55.3 & 56.3 & 57.6 & 58.3 & 62.8 & \textbf{67.3} & 62.5 \\ \hline
	\end{tabular}
}
\end{table}

\subsection{Evaluation of Dynamic Distribution Adaptation}

\begin{figure*}[t!]
	\centering
	\subfigure[Transfer results of different $\mu$]{
		\includegraphics[scale=0.45]{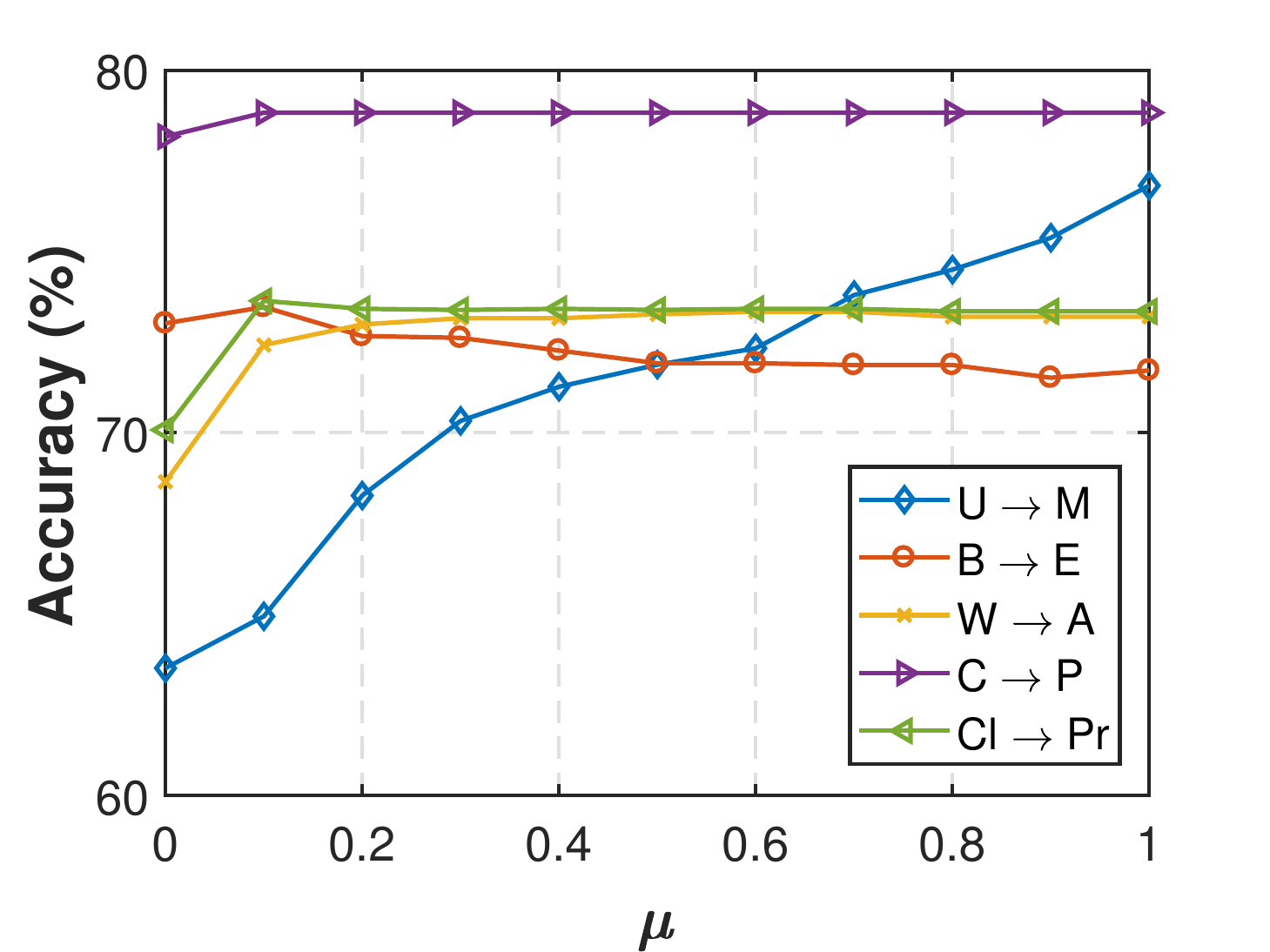}
		\label{fig-sub-mu}}
	\subfigure[Comparison of estimating $\mu$]{
		\includegraphics[scale=0.45]{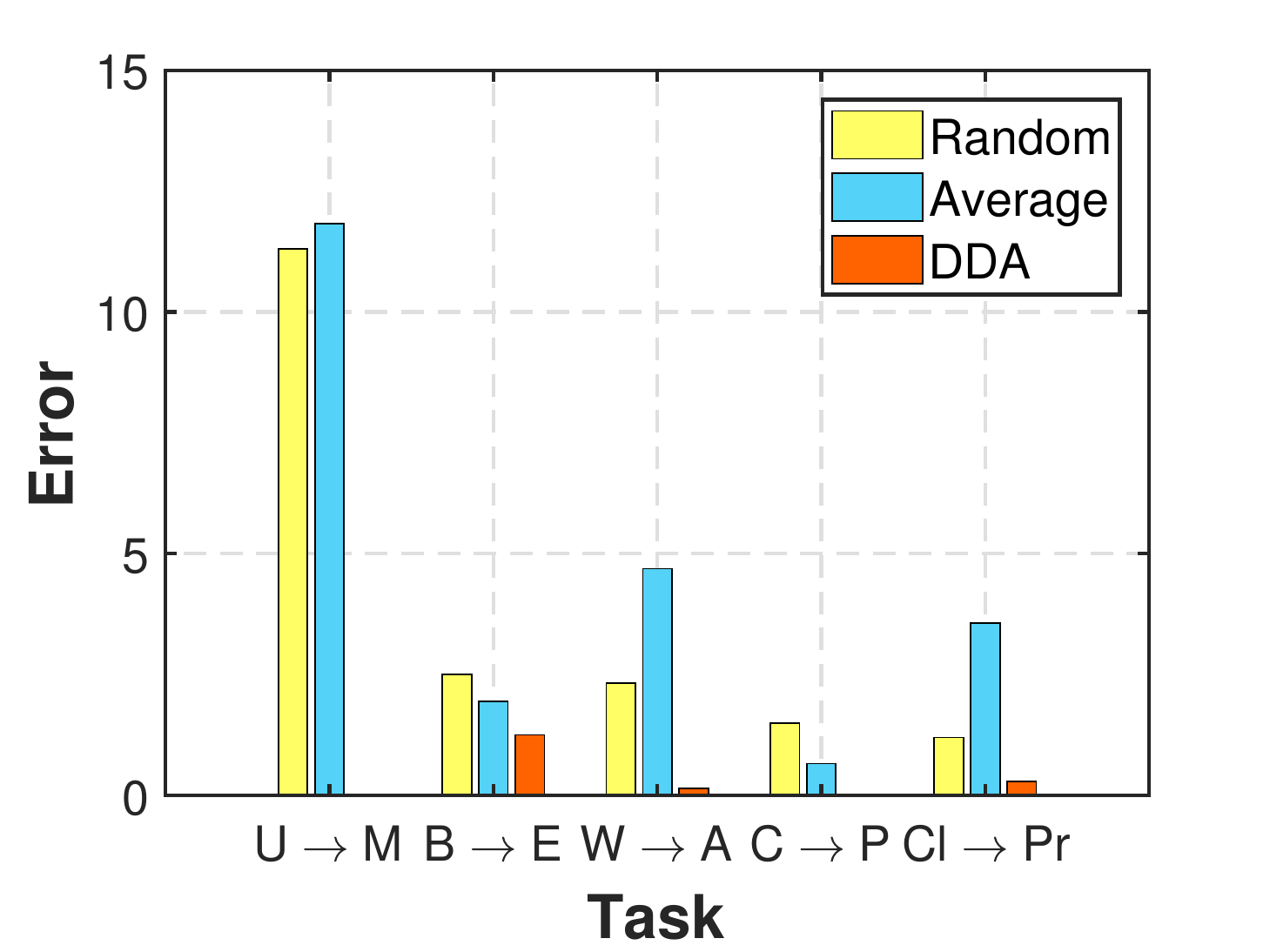}
		\label{fig-sub-mutask}}
	\vspace{-.2in}
	\caption{(a)~Performance of several tasks when searching $\mu$ in $[0,1]$. (b)~Performance comparison of Random guessing (Random), average search (AVSE), and our DDA.}
	\label{fig-mu2}
	\vspace{-.2in}
\end{figure*}

We verify the effectiveness of dynamic distribution adaptation in this section. We answer two important questions: 1) Does the different effect of marginal and conditional distributions exist in transfer learning? And 2) Is our evaluation algorithm for DDA effective? It is worth noting that there is no ground-truth for $\mu$. Therefore, in order to verify our evaluation, we record the performance of DDA by searching different $\mu$ values. The hint is that better $\mu$ value contributes better performance. Specifically, we run DDA by searching $\mu \in \{0,0.1,\cdots,0.9,1.0\}$. To answer the first question, we draw the results of DDA under different values of $\mu$ in Fig.~\ref{fig-sub-mu}. To answer the second question, we compare the error made by random search (Random), average search (Average), and our evaluation in Fig.~\ref{fig-sub-mutask}.  

Firstly, it is clear that the classification accuracy varies with different choices of $\mu$. This indicates the \textit{necessity} to consider the different effects between marginal and conditional distributions. We can also observe that the optimal $\mu$ value varies on different tasks ($\mu=0.2,0,1$ for the three tasks, respectively). Thus, it is necessary to dynamically adjust the distribution alignment between domains according to different tasks. Moreover, the optimal value of $\mu$ is \textit{not} unique for a given task. The classification results may be the same even for different values of $\mu$.

Secondly, we also report the results of our evaluation and average search in Table~\ref{tb-mu}. Combining the results in Fig.~\ref{fig-sub-mutask}, we can conclude that our evaluation of $\mu$ is significantly better than random search and average search. Additionally, both random search and average search require to run the whole MDDA or DDAN algorithm several times to get steady results, our evaluation is only required once in each iteration of the algorithm. This means that our evaluation is more efficient. It is worth noting that our evaluation is extremely close to the results from grid search. Note that on task M $\rightarrow$ U of Table~\ref{tb-mu}, our evaluation exceeds the results of grid search. Considering that there is often few or none labels in the target domain, grid search is not actually possible. Therefore, our evaluation of $\mu$ can be used to approximate the ground truth in real applications. 

Thirdly, we noticed that on image classification datasets, the performance of MDDA is slightly better than DDAN. MDDA is a shallow learning method, which is much easier to tune hyperparameters than DDAN, which is based on deep learning. We think that after a more extensive hyperparameter tuning process, the performance of DDAN will be the same or better as MDDA.

\begin{table}[t!]
	\caption{Comparison of the performance between our evaluation of $\mu$ and average search (AVSE). Suppose the results of grid search are 0.}
	\label{tb-mu}
	\resizebox{1\textwidth}{!}{
	\begin{tabular}{|c|c|c|c|c|c|c|c|c|c|c||c|}
		\hline
		Task     & M $\rightarrow$ U & B $\rightarrow$ E & E $\rightarrow$ D & A $\rightarrow$ W & W $\rightarrow$ A & C $\rightarrow$ P & P $\rightarrow$ C & Ar $\rightarrow$ Rw & Cl $\rightarrow$ Pr & Rw $\rightarrow$ Cl & AVG  \\ \hline
		AVSE  & -11.83 & -1.95  & -1.40  & -4.03  & -4.69  & -0.66  & -0.67  & -1.49    & -3.56    & -1.21  & -3.15 \\ \hline
		Ours & \textbf{\underline{+0.22}} & -1.25  & -0.80  & -1.26  & -0.14  & -0.00  & -0.00  & -0.02    & -0.29    & -0.02  & \textbf{-0.37} \\ \hline
	\end{tabular}
}
\vspace{-.1in}
\end{table}

\begin{figure*}[t!]
	\centering
	\hspace{-.1in}
	\subfigure[Ablation study of MDDA]{
		\includegraphics[scale=0.44]{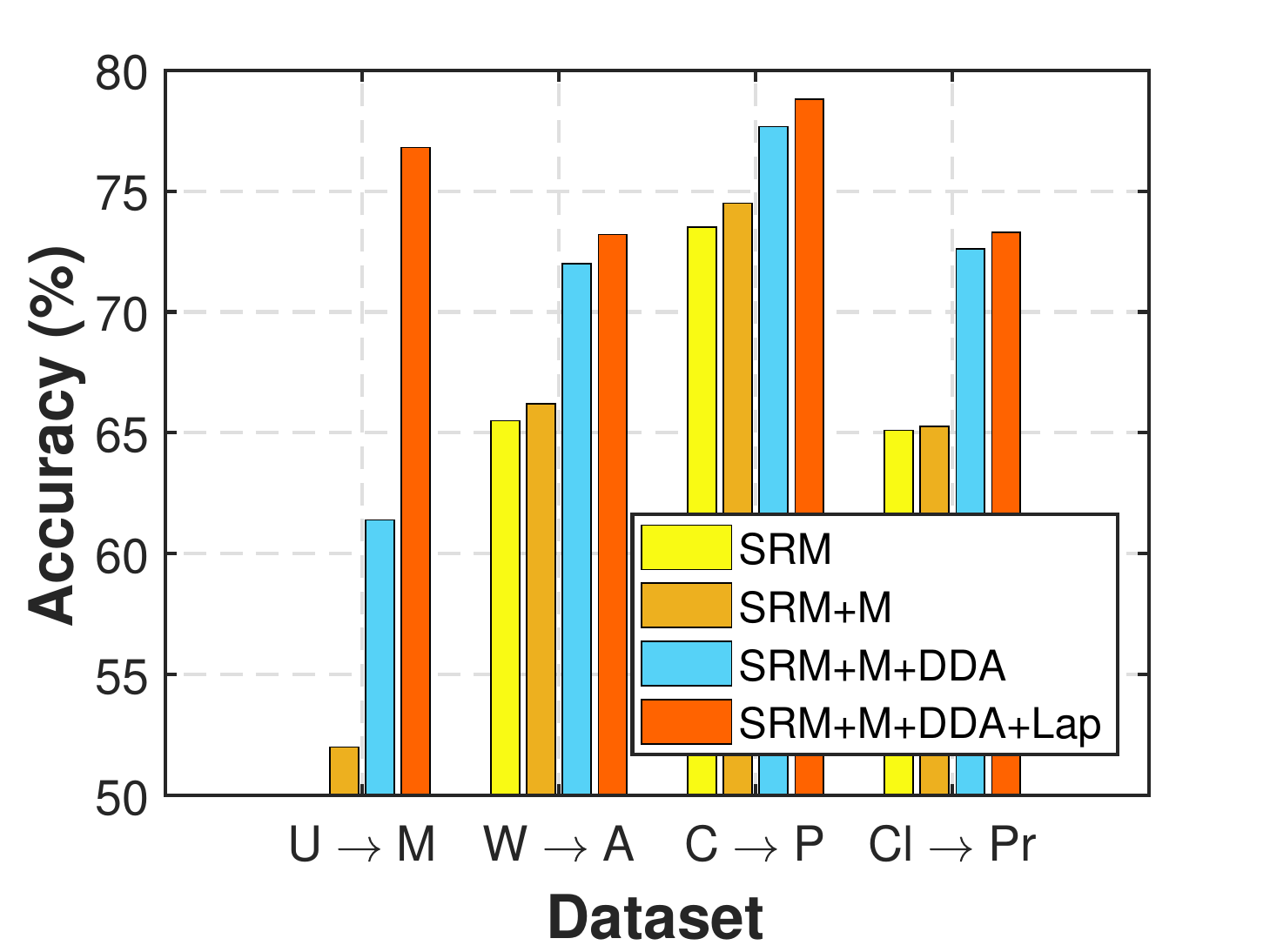}
		\label{fig-ablation-manifold}}
	\hspace{-.1in}
	\subfigure[Ablation study of DDAN]{
		\includegraphics[scale=0.44]{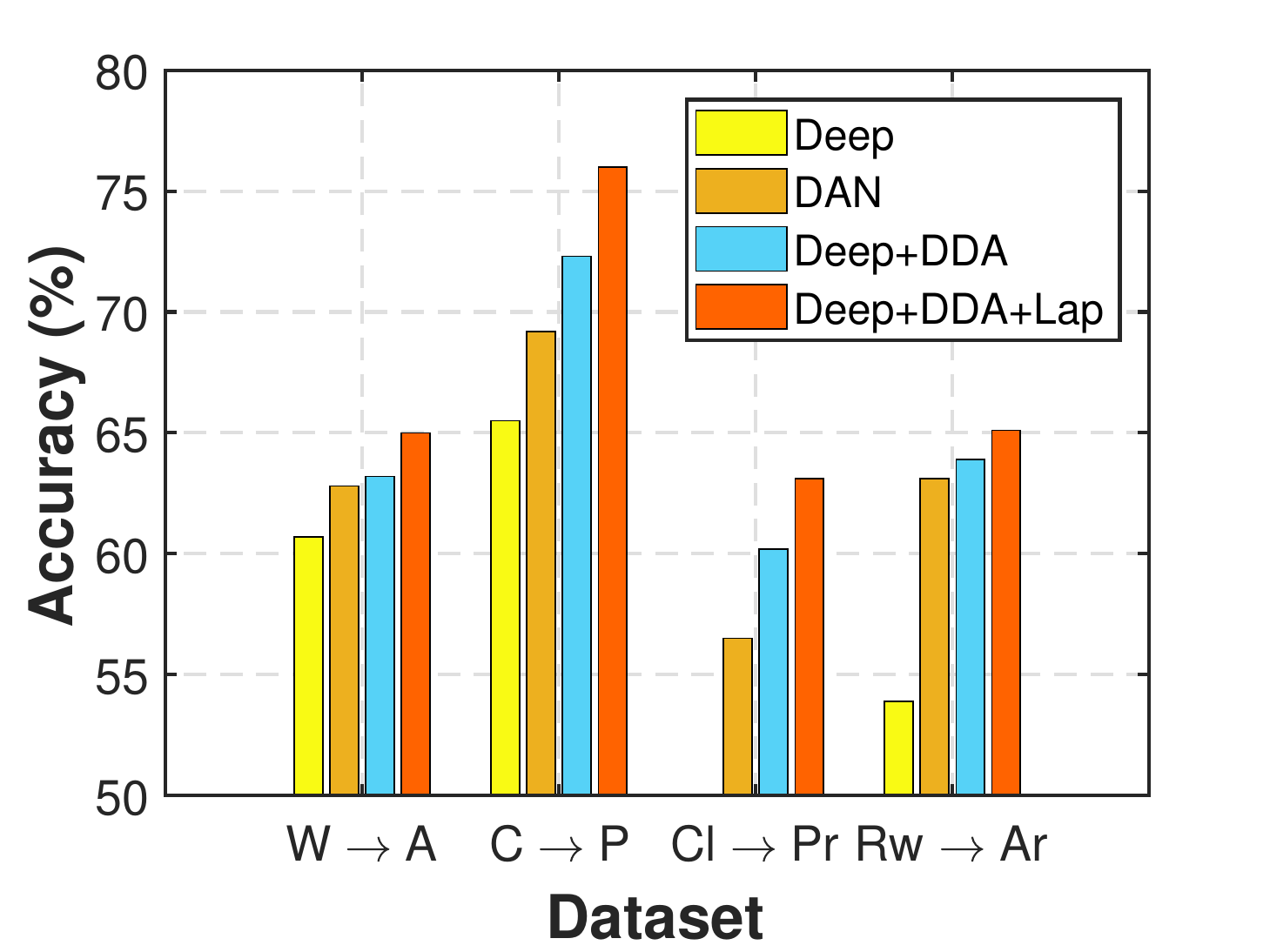}
		\label{fig-ablation-deep}}
	\vspace{-.2in}
	\caption{Ablation study of MDDA and DDAN. `M' denotes manifold learning, and `Lap' denotes Laplace regularization.}
	\label{fig-ablation}
	\vspace{-.2in}
\end{figure*}

\subsection{Ablation Study}

In this section, we conduct ablation study of MDDA and DDAN. MDDA mainly consists of four components: SRM, manifold learning, DDA, and Laplacian regularization. DDAN is composed of a deep network, DDA, and Laplacian regularization. We extensively analyze the performance of MDDA and DDAN on some tasks from each dataset and present the results in Fig.~\ref{fig-ablation}. 

The results clearly indicate that each component is important to DDA. Of all the components, it is shown that our proposed DDA component is the most important part, which dramatically increases the results of transfer learning. For MDDA, manifold feature learning shows marginal improvement, while it could help to eliminate the feature distortion of the original space~\cite{baktashmotlagh2013unsupervised}. For DDAN, we can clearly see that our DDA component is better than DAN, which is only adapting the marginal distributions. This again clarifies the importance of the proposed DDA framework. Finally, It seems that Laplacian regularization also generates marginal improvements except on the digit datasets (USPS+MNIST). We add Laplacian regularization since it helps the algorithm to converge quickly.

\begin{figure*}[t!]
	\centering
	\subfigure[$d$]{
		\includegraphics[scale=0.4]{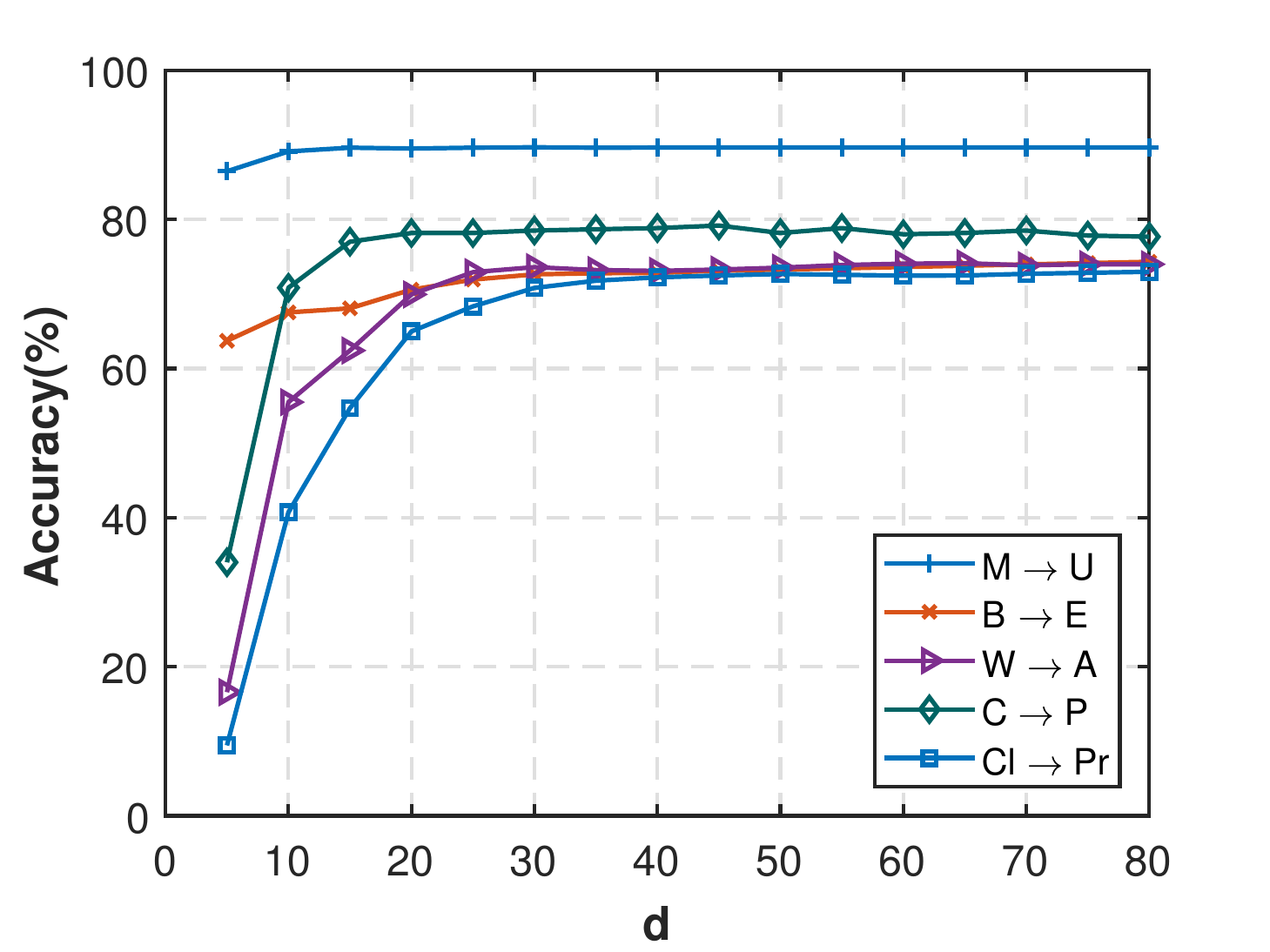}
		\label{fig-sub-d}}
	\hspace{-.1in}
	\subfigure[$\lambda$]{
		\includegraphics[scale=0.4]{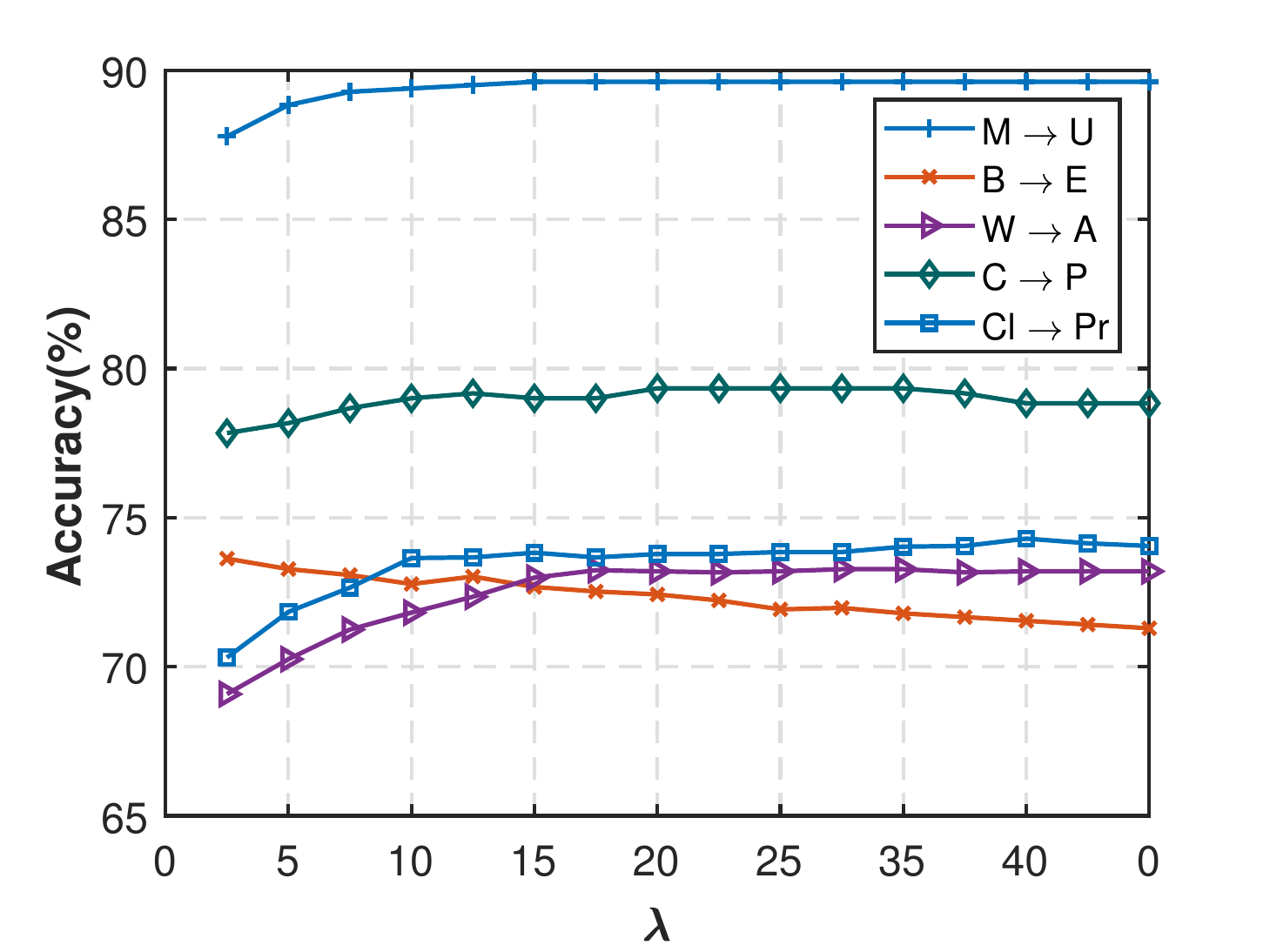}
		\label{fig-sub-p}}
	\hspace{-.1in}
	\subfigure[$p$]{
		\includegraphics[scale=0.4]{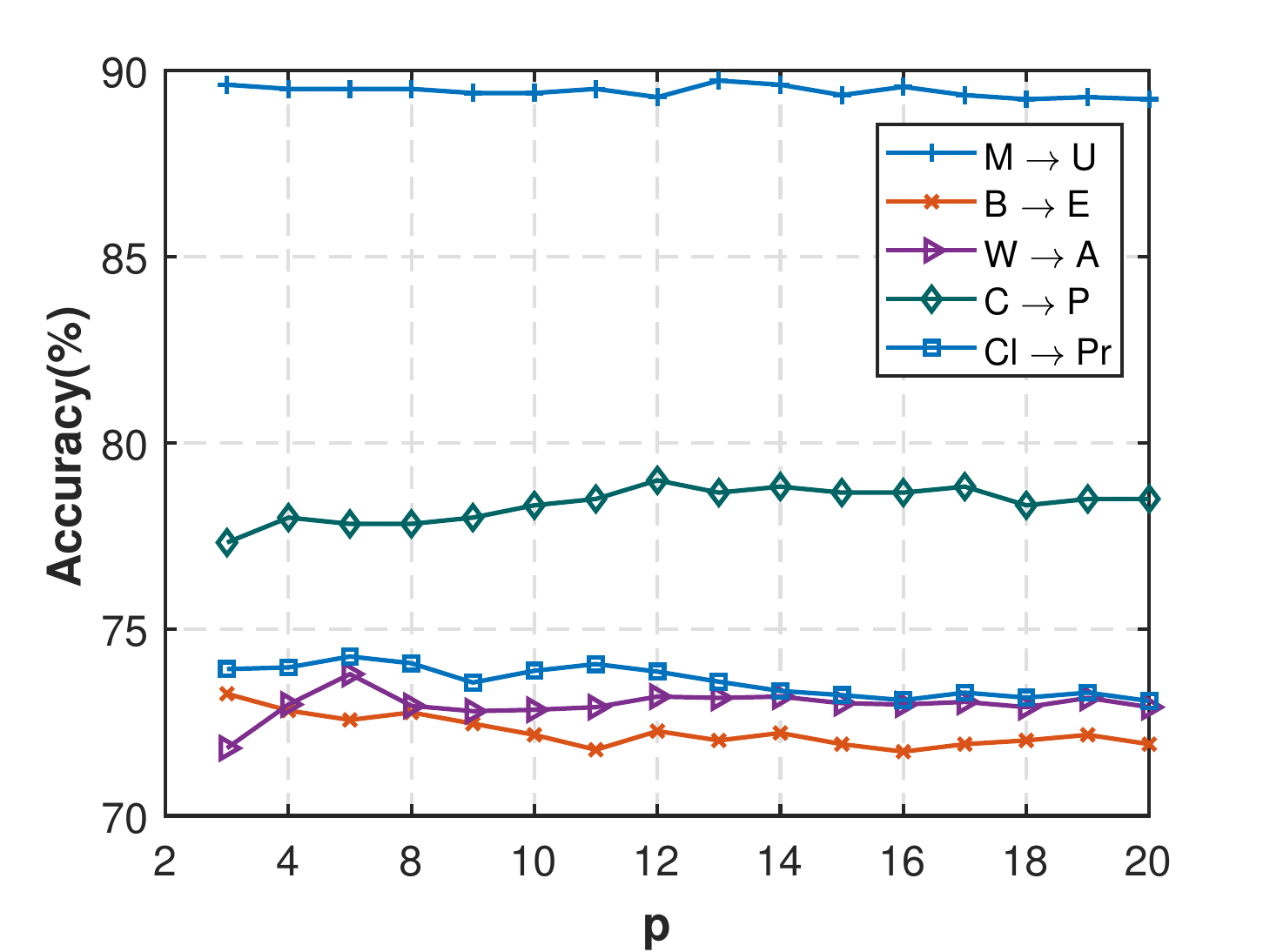}
		\label{fig-sub-lambda}}
	\hspace{-.1in}
	\subfigure[$\eta$]{
		\includegraphics[scale=0.4]{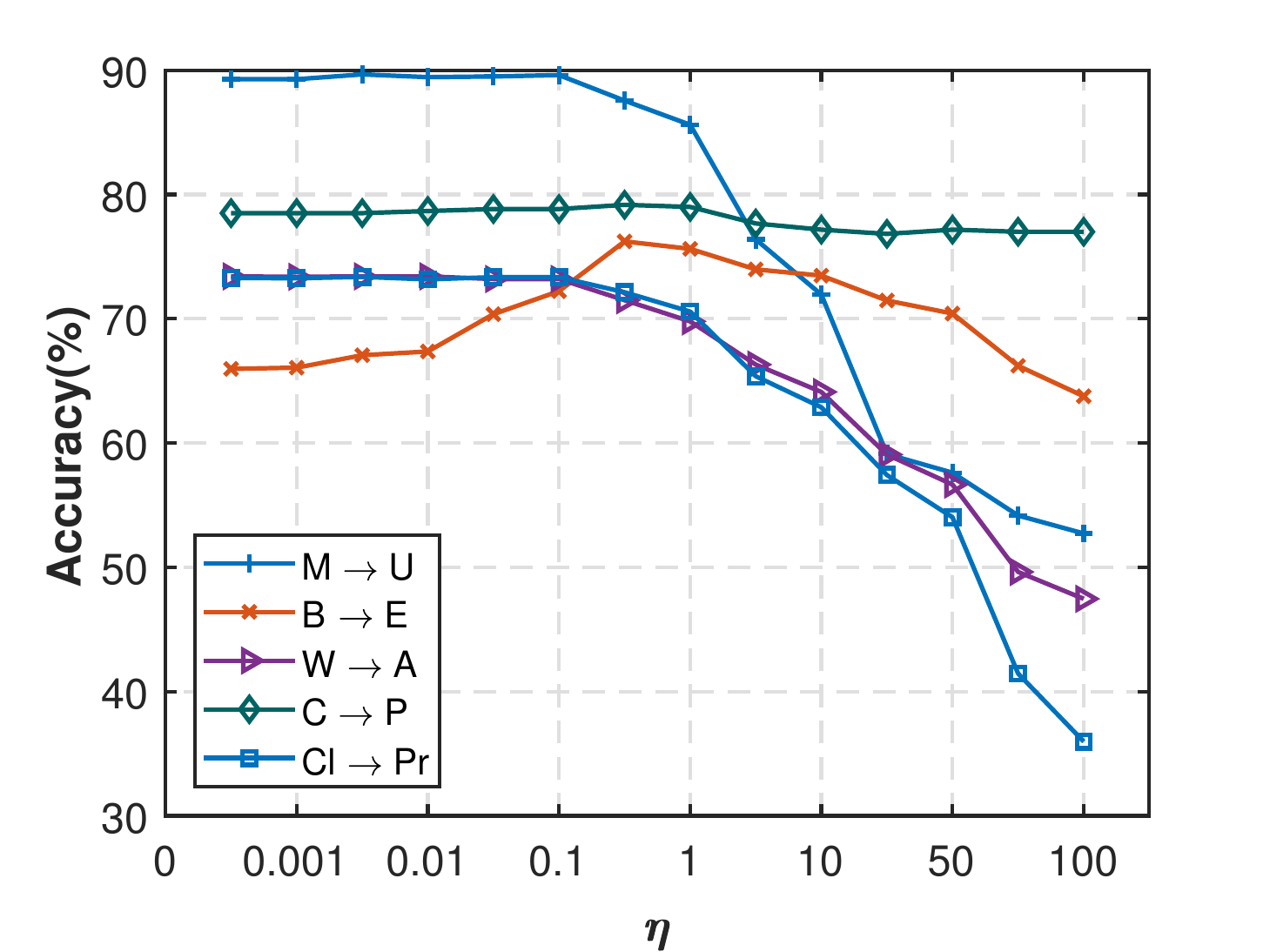}
		\label{fig-sub-eta}}
	\hspace{-.1in}
	\subfigure[$\rho$]{
		\includegraphics[scale=0.4]{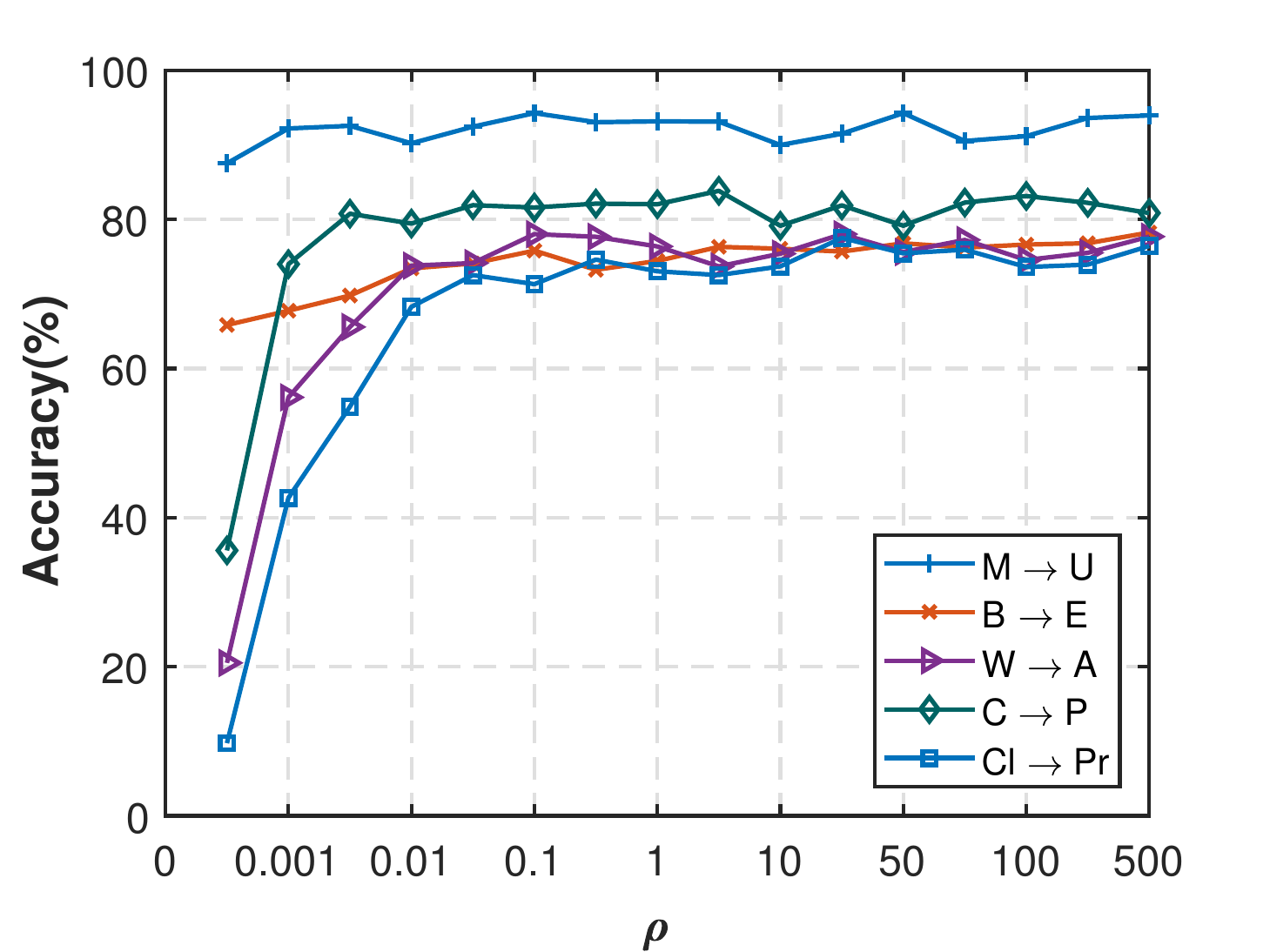}
		\label{fig-sub-rho}}
	\hspace{-.1in}
	\subfigure[Convergence]{
		\includegraphics[scale=0.4]{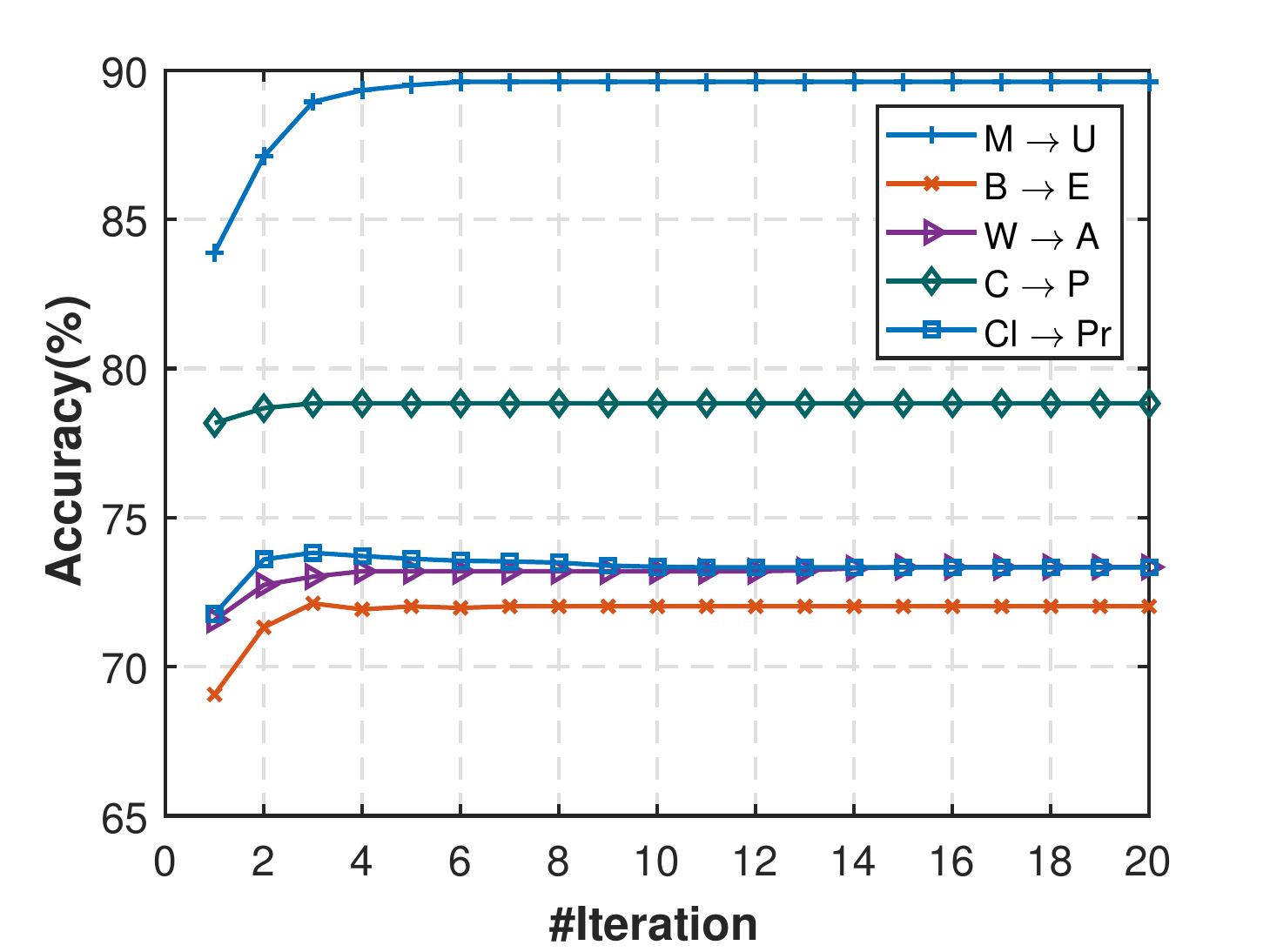}
		\label{fig-sub-iteration}}
	\vspace{-.2in}
	\caption{Parameter sensitivity analysis and convergence of MDDA.}
	\label{fig-p-d}

\end{figure*}

%


\subsection{Parameter Sensitivity and Convergence Analysis}
\label{sec-para}

As with other state-of-the-art domain adaptation algorithms~\cite{zhang2017joint,long2014adaptation,ghifary2017scatter}, MDDA and DDAN also involve several parameters. In this section, we evaluate the parameter sensitivity of them. Experimental results demonstrated the robustness of MDDA and DDAN under a wide range of parameter choices.

\begin{figure*}[t!]
	\centering
	
	\subfigure[$\lambda$]{
		\includegraphics[scale=0.4]{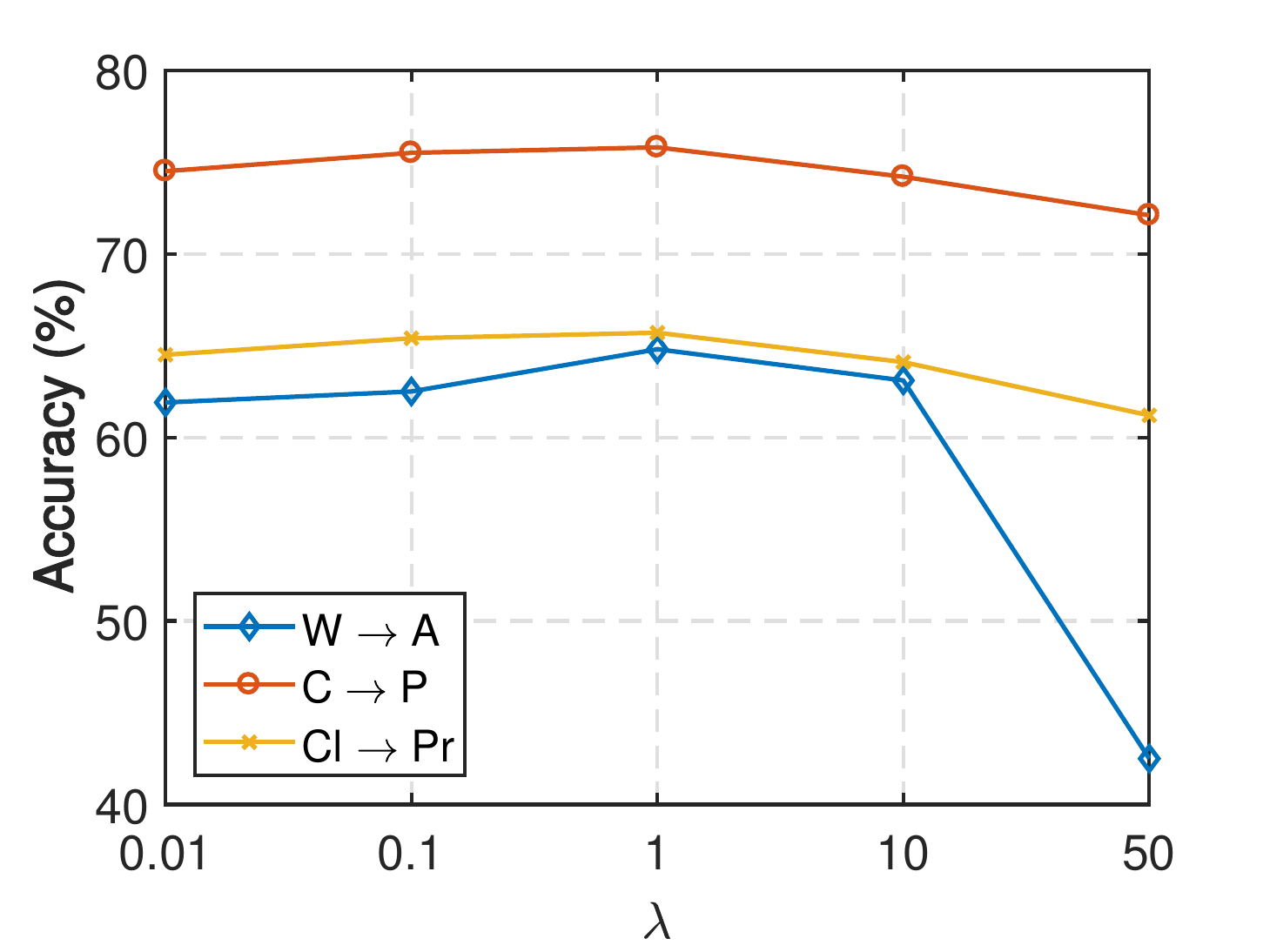}
		\label{fig-sub-ddan-lambda}}
	\hspace{-.1in}
	\subfigure[$\rho$]{
		\includegraphics[scale=0.4]{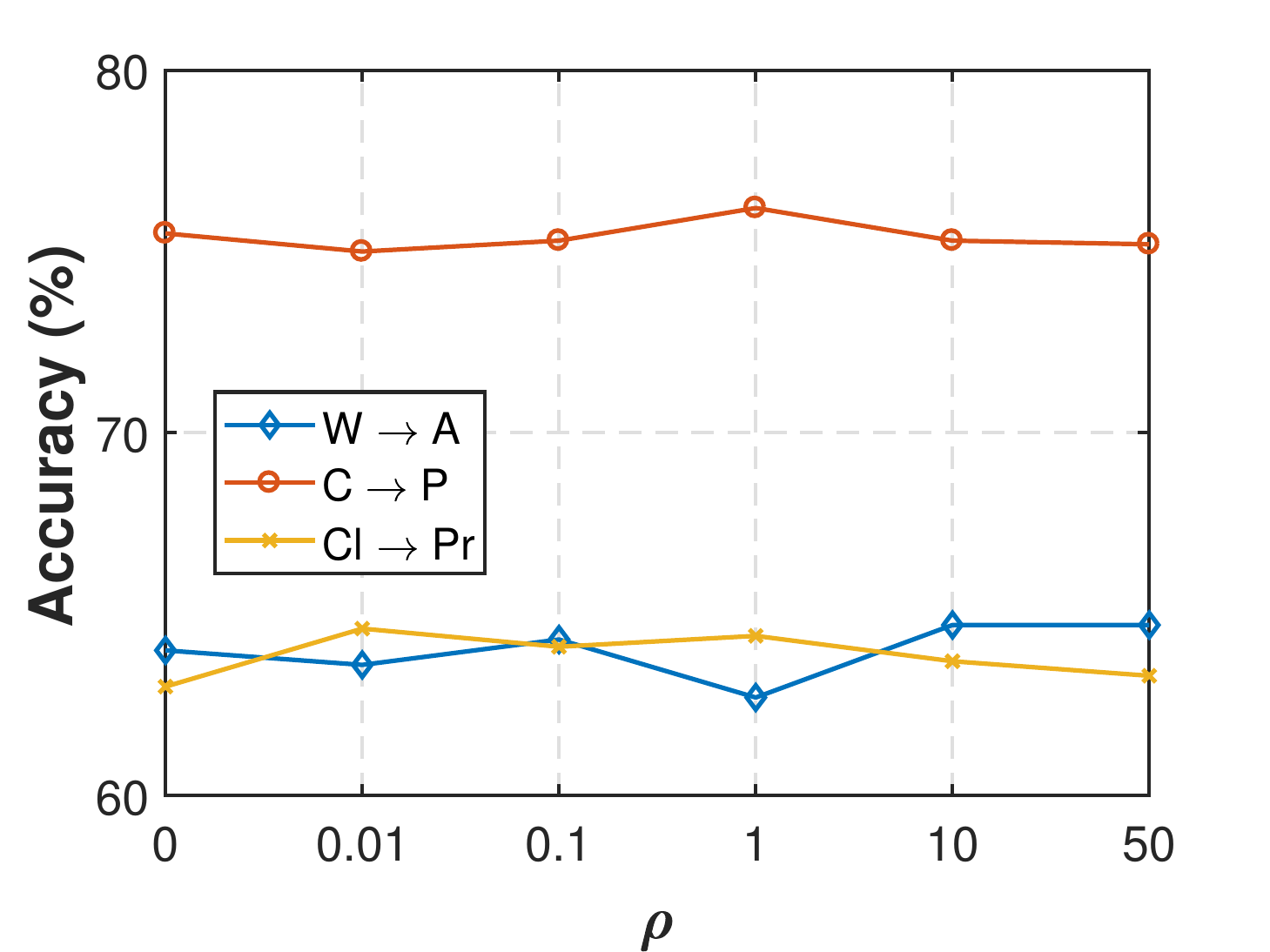}
		\label{fig-sub-ddan-p}}
	\hspace{-.2in}
	\subfigure[$p$]{
		\includegraphics[scale=0.4]{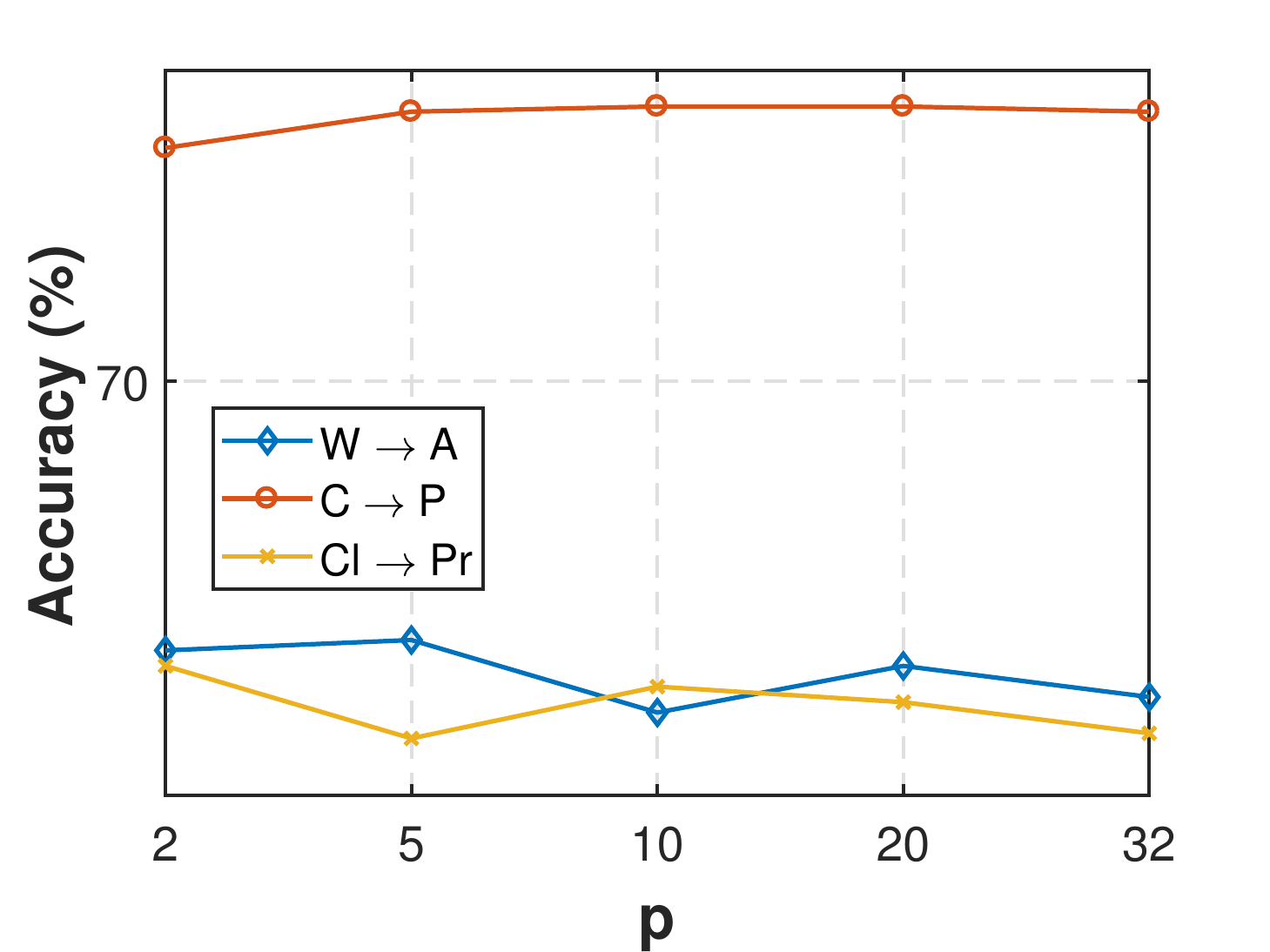}
		\label{fig-sub-ddan-rho}}
	\hspace{-.1in}
	\subfigure[Convergence]{
		\includegraphics[scale=0.4]{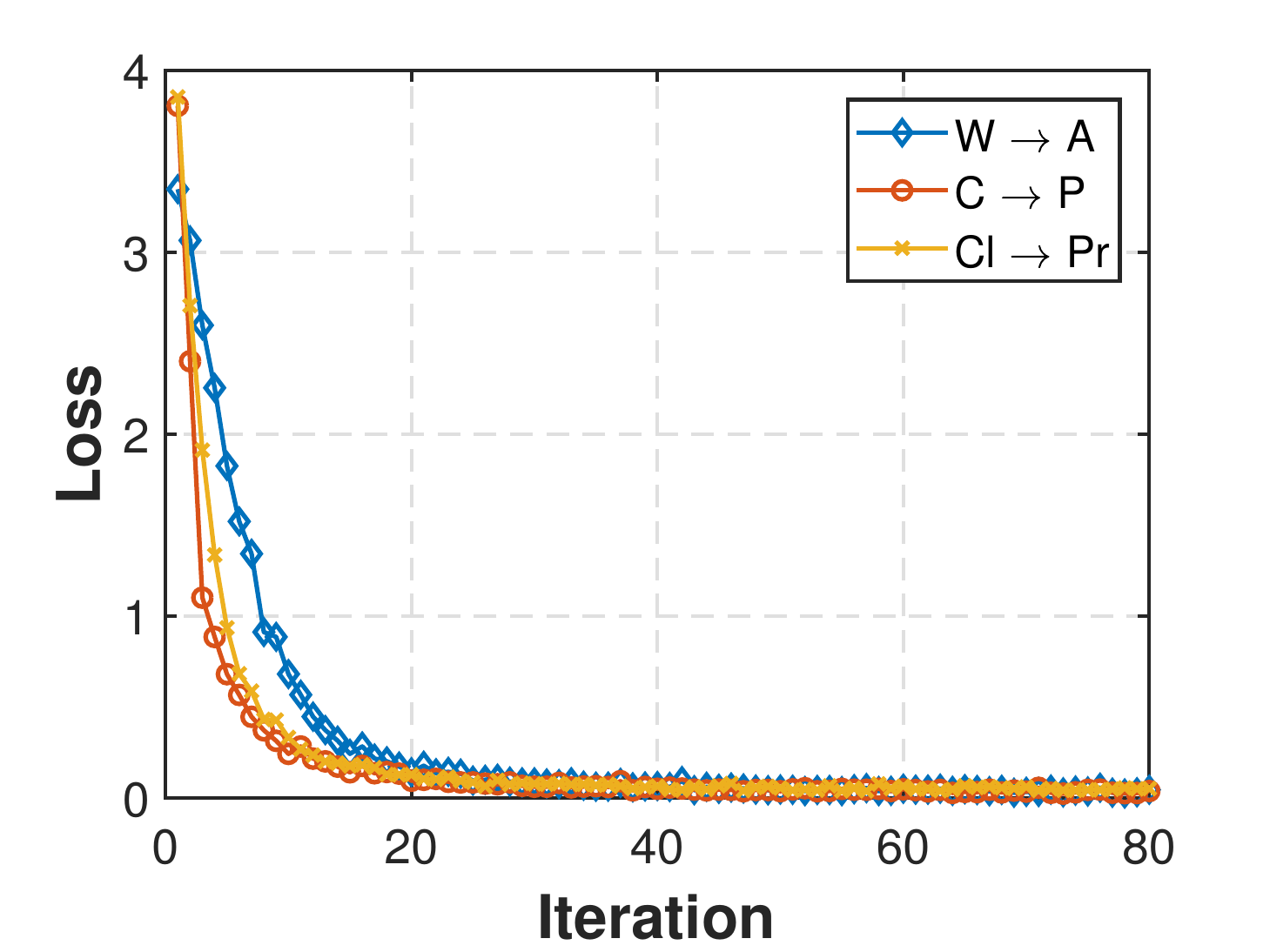}
		\label{fig-sub-ddan-iteration}}
	\vspace{-.2in}
	\caption{Parameter sensitivity and convergence analysis of DDAN.}
	\label{fig-ddan-para}

\end{figure*}

\begin{table}[t!]
\caption{Running time of MDDA and DDAN}
\vspace{-.1in}
\label{tb-time}
\begin{tabular}{|c|c|c|c|c|c|c|}
\hline
Task & ARTL & JGSA & DANN & CDAN & MDDA & DDAN \\ \hline \hline
U $\rightarrow$ M & 29.1 & 14.6 & - & - & 31.4 & - \\ \hline
B $\rightarrow$ E & 22.8 & 18.7 & - & - & 23.5 & - \\ \hline
W $\rightarrow$ A & 45.6+763.6 & 66.5 + 763.6 & 1567.3 & 1873.2 & 48.8 + 7663.6 & 1324.1 \\ \hline
C $\rightarrow$ P & 124.2 + 1321.4 & 198.3 + 1321.4 & 2342.1 & 2451.2 & 156.7 + 1321.4 & 2109.8 \\ \hline
Cl $\rightarrow$ Pr & 187.4 + 1768.7 & 244.3 + 1768.7 & 2877.7 & 2956.5 & 207.4 + 1768.7 & 2698.1 \\ \hline
\end{tabular}
\end{table}

\subsubsection{MDDA}

We investigated the sensitivity against manifold subspace dimension $d$ and \#neighbor $p$ through experiments with a wide range of $d \in \{10,20,\cdots,100\}$ and $p \in \{2,4,\cdots,64\}$ on randomly selected tasks. From the results in Fig.~\ref{fig-sub-d} and \ref{fig-sub-p}, it can be observed that MDDA is robust with regard to different values of $d$ and $p$. Therefore, they can be selected without in-depth knowledge of specific applications. 

We ran MDDA with a wide range of values for regularization parameters $\lambda,\eta$, and $\rho$ on several random tasks and compare its performance with the best baseline method. We only report the results of $\lambda$ in Fig.~\ref{fig-sub-lambda}, and the results of $\rho$ and $\eta$ are following the same tendency. We observed that MDDA can achieve a robust performance with regard to a wide range of parameter values. Specifically, the best choices of these parameters are: $\lambda \in [0.5, 1,000], \eta \in [0.01,1]$, and $\rho \in [0.01,5]$.

We evaluate the convergence of MDDA through experimental analysis. From the results in Fig.~\ref{fig-sub-iteration}, it can be observed that MDDA can reach a steady performance in only a few $(T < 10)$ iterations. It indicates the training advantage of MDDA in cross-domain tasks.

\subsubsection{DDAN}

DDAN involves three key parameters: $\lambda, p,$ and $ \rho$. Similar to MDDA, we report the parameter sensitivity and convergence in Fig.~\ref{fig-ddan-para}. It is clear that DDAN is robust to these parameters. Therefore, in real applications, the hyperparameters of DDAN do not have to be cherry-picked. This is extremely important in deep learning since it is rather time-consuming to tune the hyperparameters.

We also extensively evaluate the convergence of DDAN in Fig.~\ref{fig-sub-ddan-iteration}. It is shown that DDAN is able to converge quickly with steady performance.

\subsubsection{Time complexity}

We empirically check the running time of MDDA and DDAN, and present the results in Table~\ref{tb-time}. Note that for image classification tasks, the running time of ARTL, JGSA, and MDDA are the summation of deep feature extraction and algorithm running time, since these algorithms require to extract features before transfer learning. It is shown that both MDDA and DDAN can achieve efficient computing time while achieving better performance compared to these comparison methods.

\section{Conclusions and Future Work}
\label{sec-con}

In this paper, to solve the transfer learning problem, we propose the novel dynamic distribution adaptation (DDA) concept. DDA is able to dynamically evaluate the relative importance between the source and target domains. Based on DDA, we propose two novel methods: the manifold DDA (MDDA) for traditional transfer learning, and deep DDA networks (DDAN) for deep transfer learning. Extensive experiments on digit recognition, sentiment analysis, and image classification have demonstrated that both MDDA and DDAN could achieve the best performance compared to other state-of-the-art traditional and deep transfer learning methods. 

In the future, we plan to extend the DDA framework into the heterogeneous transfer learning areas as well as apply it to more complex transfer learning situations.


\section*{Acknowledgments}

This work is supported in part by National Key R \& D Program of China (2016YFB1001200), NSFC (61572471, 61972383), the support of grants from Hong Kong CERG projects 16209715, 16244616, Nanyang Technological University, Nanyang Assistant Professorship (NAP), and Beijing Municipal Science \& Technology Commission~(Z171100000117017).

%
\bibliographystyle{ACM-Reference-Format}
\bibliography{tist19}

\end{document}